\documentclass[letterpaper]{article} % DO NOT CHANGE THIS
\usepackage{aaai2026}  % DO NOT CHANGE THIS
\usepackage{times}  % DO NOT CHANGE THIS
\usepackage{helvet}  % DO NOT CHANGE THIS
\usepackage{courier}  % DO NOT CHANGE THIS
\usepackage[hyphens]{url}  % DO NOT CHANGE THIS
\usepackage{graphicx} % DO NOT CHANGE THIS
\usepackage{natbib}  % DO NOT CHANGE THIS AND DO NOT ADD ANY OPTIONS TO IT
\usepackage{caption} % DO NOT CHANGE THIS AND DO NOT ADD ANY OPTIONS TO IT
\usepackage{multirow}
\usepackage{amsmath}
\usepackage{amsfonts}
\usepackage{amssymb}
\usepackage{booktabs}
\usepackage{cuted}
\usepackage{algorithm}
\usepackage{algorithmic}

\usepackage{newfloat}
\usepackage{listings}
\DeclareCaptionStyle{ruled}{labelfont=normalfont,labelsep=colon,strut=off} % DO NOT CHANGE THIS
\floatstyle{ruled}
\newfloat{listing}{tb}{lst}{}
\floatname{listing}{Listing}
\title{IAD-R1: Reinforcing Consistent Reasoning in Industrial Anomaly Detection}
\author{
    Yanhui Li\textsuperscript{\rm 1},
    Yunkang Cao\textsuperscript{\rm 2},
    Chengliang Liu\textsuperscript{\rm 3},
    Yuan Xiong\textsuperscript{\rm 1},
    Xinghui Dong\textsuperscript{\rm 4},
    Chao Huang\textsuperscript{\rm 1}\thanks{Corresponding author}
}
\affiliations{
    \textsuperscript{\rm 1}	Sun Yat-sen University
    \textsuperscript{\rm 2} Hunan University
    \\
    \textsuperscript{\rm 3} University of Macau
    \textsuperscript{\rm 4} Ocean University of China
    liyh665@mail2.sysu.edu.cn, caoyunkang0207@gmail.com, liucl1996@163.com \\
    \{xiongy89, huangch253\}@mail.sysu.edu.cn, xinghui.dong@ouc.edu.cn
}

\usepackage{bibentry}
\begin{document}

\maketitle

\begin{strip}
    \vspace{-10pt}
    \centering\\[-30pt]
    \includegraphics[width=1\textwidth]{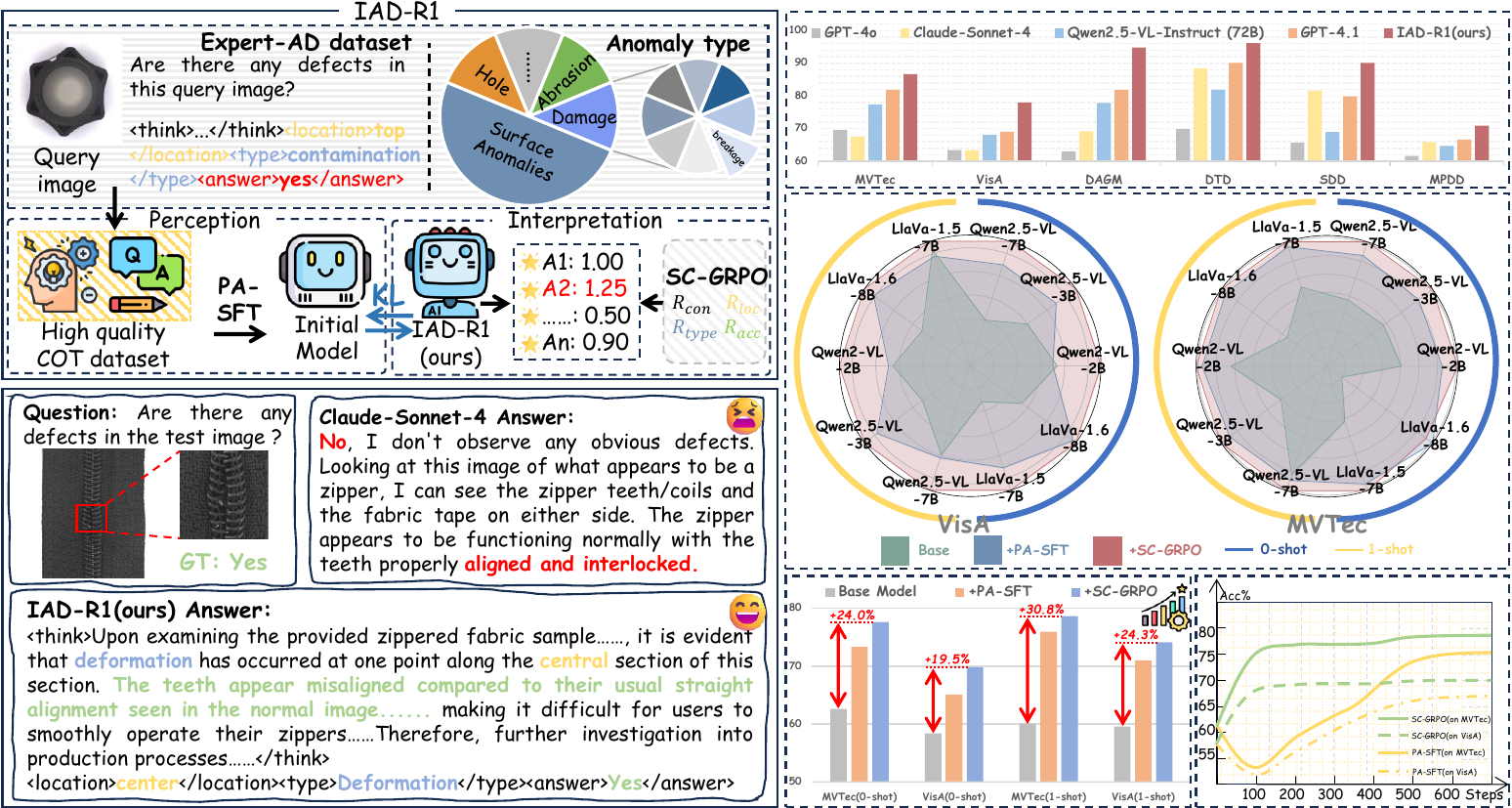}
    \captionof{figure}{Overview of IAD-R1. The top left panel illustrates the composition of the Expert-AD and the two-stage training framework of IAD-R1. The bottom left panel shows an output example of IAD-R1 for anomaly detection. The right panels present quantitative analyses showcasing the performance of IAD-R1 across different model configurations and datasets.}
    \label{overview}
\end{strip}

%摘要0_abstract
\begin{abstract}
Industrial anomaly detection is a critical component of modern manufacturing, yet the scarcity of defective samples restricts traditional detection methods to scenario-specific applications. Although Vision-Language Models (VLMs) demonstrate significant advantages in generalization capabilities, their performance in industrial anomaly detection remains limited. To address this challenge, we propose IAD-R1, a universal post-training framework applicable to VLMs of different architectures and parameter scales, which substantially enhances their anomaly detection capabilities. IAD-R1 employs a two-stage training strategy: the Perception Activation Supervised Fine-Tuning (PA-SFT) stage utilizes a meticulously constructed high-quality Chain-of-Thought dataset (Expert-AD) for training, enhancing anomaly perception capabilities and establishing reasoning-to-answer correlations; the Structured Control Group Relative Policy Optimization (SC-GRPO) stage employs carefully designed reward functions to achieve a capability leap from ``Anomaly Perception" to ``Anomaly Interpretation". Experimental results demonstrate that IAD-R1 achieves significant improvements across 7 VLMs, the largest improvement was on the DAGM dataset, with average accuracy 43.3\% higher than the 0.5B baseline. Notably, the 0.5B parameter model trained with IAD-R1 surpasses commercial models including GPT-4.1 and Claude-Sonnet-4 in zero-shot settings, demonstrating the effectiveness and superiority of IAD-R1. The dataset, code, and all model weights will be publicly available at \url{https://github.com/Yanhui-Lee/IAD-R1}.

\end{abstract}
%简介1_introduction
\section{Introduction}
Industrial anomaly detection serves as a critical component of modern manufacturing quality control, facing challenges including diverse anomaly types, large inter-class variance, and scarce anomaly samples~\cite{pni,iad-survey,patchcore,winclip}. The complexity and diversity of anomaly patterns make building universal detection models extremely challenging. Traditional methods primarily rely on hand-crafted feature extractors and domain-specific expert knowledge~\cite{uniad,diad,vpdm,inctrl}, but these approaches are often limited to single product categories, severely lacking generalization capabilities.

In recent years, Vision-Language Models (VLMs) have provided new possibilities for solving industrial anomaly detection problems through their powerful multimodal understanding and generalization capabilities~\cite{lad-reasoner,logicad,triad,lr-iad,mmad}. Researchers have primarily adopted two strategies to apply VLMs to anomaly detection: (1) Using traditional anomaly detection methods as visual anomaly experts, feeding their generated anomaly localization results along with prompt text and test images into VLMs for judgment~\cite{anomalygpt,myriad,anomaly-ov,echo}; (2) Based on question-answering anomaly detection datasets, directly adapting VLMs to industrial anomaly detection tasks through post-training techniques such as supervised fine-tuning or reinforcement learning~\cite{anomalyr1,lr-iad,triad}.

However, both strategies suffer from fundamental limitations. The anomaly expert-assisted approach has overall performance constrained by the capability ceiling of the selected anomaly expert, making it difficult to breakthrough the bottlenecks of traditional methods. End-to-end fine-tuning methods, while avoiding dependence on traditional algorithms, face deeper issues: existing training data lacks high-quality reasoning process annotations, leading models to learn only simple input-output mappings without mastering the intrinsic logic of anomaly analysis. More critically, traditional supervised fine-tuning easily leads to cognitive rigidity, while existing reinforcement learning methods, due to coarse reward design, frequently exhibit inconsistency between reasoning processes and final answers.

Successful experiences in general domains~\cite{deepseek-r1,vlm-r1} demonstrate that high-quality chain-of-thought (CoT)~\cite{cot} data and reinforcement learning strategies can improve model reasoning capabilities. However, the exploration of the R1-style methods in the anomaly detection domain remains insufficient, lacking high-quality CoT datasets for this task and specifically designed reinforcement learning objectives.

To address these issues, we propose IAD-R1, a two-stage post-training framework specifically designed for industrial anomaly detection, aimed at addressing the scarcity of high-quality CoT data and the lack of reasoning-answer consistency in existing models. As shown in Figure \ref{overview}, we construct the Expert-AD (totaling 5.9K QA pairs) dataset, the first industrial anomaly detection dataset containing high-quality CoT reasoning. Its CoT template follows a progressive three-layer approach of spatial perception, knowledge-driven analysis, and comprehensive decision-making to systematically simulate the expert anomaly detection process. Inspired by successful two-stage training experiences in DeepSeek-R1~\cite{deepseek-r1}, we propose the Perception Activation Supervised Fine-Tuning Strategy (PA-SFT) and Structured Control Group Relative Policy Optimization (SC-GRPO) strategies. Targeting the specificity of industrial anomaly detection, we design multi-dimensional reward functions: $R_{acc}$, $R_{loc}$, $R_{type}$, and $R_{con}$ to achieve refined optimization of model reasoning processes, thereby enhancing the model's anomaly understanding capabilities.

Extensive experimental results demonstrate that IAD-R1 can significantly improve anomaly detection performance across different backbones, achieving up to 43.3\% average accuracy. Notably, small parameter models (0.5B) trained with IAD-R1 not only outperform larger baseline models (72B) in zero-shot settings but also surpass advanced commercial models such as GPT-4.1~\cite{gpt4.1} and Claude-Sonnet-4~\cite{claude4}, validating the superior parameter efficiency and performance advantages of our method. IAD-R1 effectively addresses the key problem of insufficient generalization capability in traditional methods, providing an efficient and universal solution for practical applications of VLMs in industrial anomaly detection.
Our main contributions are summarized as follows:
\begin{itemize}
    \item We construct the first industrial anomaly detection dataset Expert-AD containing high-quality CoT reasoning, providing crucial data support for training.
    \item We propose IAD-R1, a two-stage post-training framework suitable for industrial anomaly detection, which guides models to achieve a capability leap from ``Anomaly Perception" to ``Anomaly Interpretation" through PA-SFT and SC-GRPO.
    \item Extensive experiments validate the effectiveness and universality of IAD-R1, providing an efficient solution for VLMs applications in industrial anomaly detection.
\end{itemize}

%相关工作2_relatedwork
\section{Related Work}
\subsection{Traditional Industrial Anomaly Detection}
Due to the challenge of scarce anomaly samples in industrial anomaly detection, researchers utilize a small number of normal samples as auxiliary datasets for training and testing. WinCLIP~\cite{winclip} pioneered the application of CLIP~\cite{clip} to industrial anomaly detection by computing the similarity between handcrafted text prompts and test images to achieve defect detection. Addressing the complexity of manual prompt design, PromptAD~\cite{promptad} simplified text prompt design through semantic connections to construct negative samples and explicit anomaly boundaries. In contrast, One-to-Normal~\cite{one-to-normal} provided visual support for detection by performing one-to-one normal transformations on query images through an anomaly-free generative model.
Considering the unpredictability of anomaly categories in production environments~\cite{anovl,aaclip,vcp-clip,bayesian,filo}, AnomalyCLIP~\cite{anomalyclip} achieved zero-shot anomaly detection by learning object-agnostic text prompts. AdaCLIP~\cite{adaclip} further enhanced anomaly detection capabilities by learning dynamic prompts for both text and images.

\subsection{VLMs on Industrial Anomaly Detection} 
With the emergence of VLMs~\cite{mmllm,mllm,mllm_survey}, researchers have begun exploring the utilization of their powerful generalization capabilities for industrial anomaly detection~\cite{triad,echo,mmad,lad-reasoner,logicad,logicqa,lr-iad}. AnomalyGPT~\cite{anomalygpt} adapted to anomaly detection tasks by generating training data through simulated anomaly images and fine-tuning the model using image decoders and prompt learners. However, due to the modality gap between textual and visual domains, Myriad~\cite{myriad} further introduced anomaly maps generated by visual experts to guide the model's attention toward anomalous regions. Addressing the requirements of zero-shot scenarios, Anomaly-OV~\cite{anomaly-ov} achieved more precise zero-shot anomaly detection through feature matching mechanisms that adaptively select and emphasize anomalous visual tokens.

%方法3_method
\section{Methodology}
\subsection{Problem Definition} In this study, we focus on zero-shot and one-shot scenarios for industrial anomaly detection, where training and testing adopt different data distribution settings. Specifically, our problem formulation is as follows. \\
\textbf{Training Phase.} We adopt a zero-shot training paradigm, where training data comes entirely from our constructed auxiliary dataset Expert-AD, containing no samples from the target test domain. \\
\textbf{Testing Phase.} We conduct evaluations under two settings on the target domain:
\begin{itemize}
    \item Zero-shot testing: The model receives only text prompts $p$ and the image to be detected $I_{test}$ as input, outputting model reasoning and anomaly detection results.
    \item One-shot testing: In addition to text prompts $p$ and test images $I_{test}$, the model additionally receives a normal reference image $I_{ref}$ from the same product category as contextual information, with the output being model reasoning and anomaly detection results.
\end{itemize}

\begin{figure*}[!t]
    \centering
    \includegraphics[width=1\linewidth]{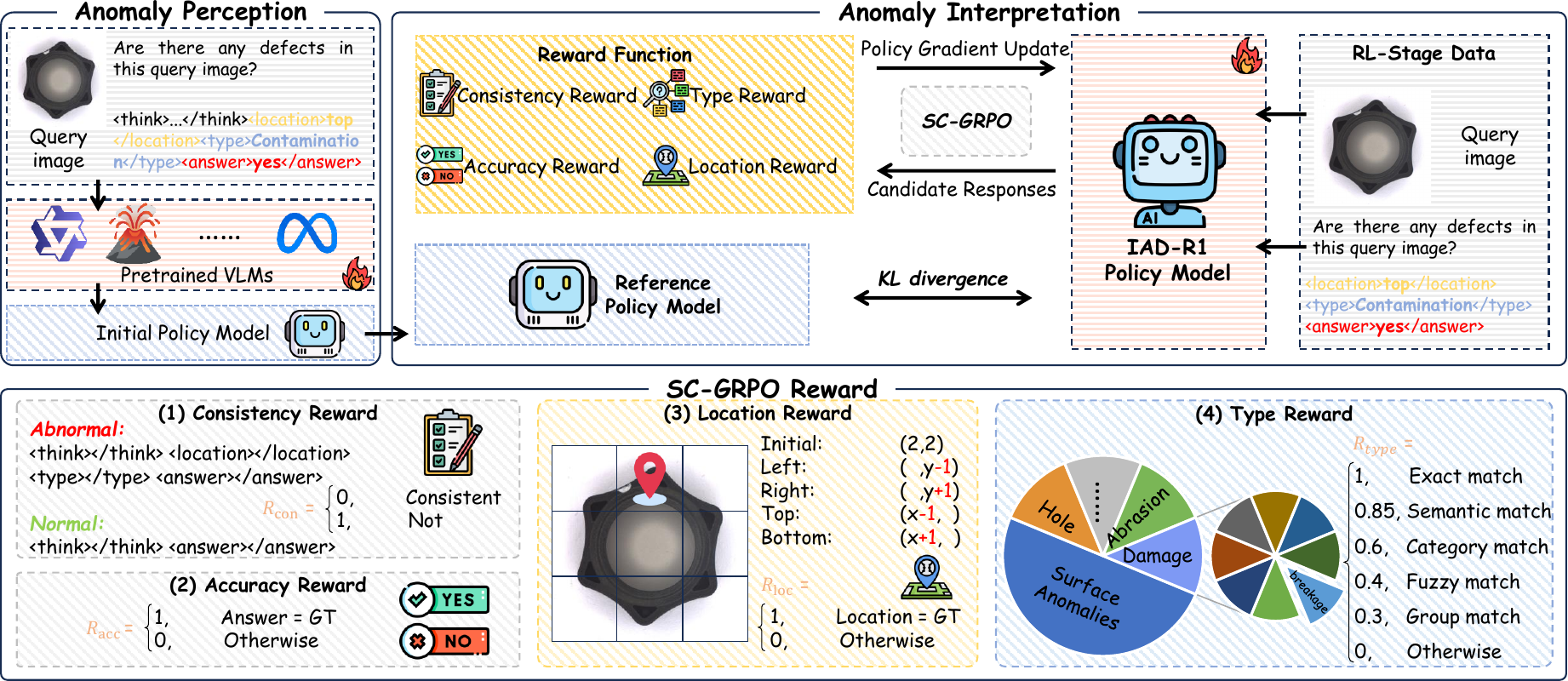}
    \caption{Architecture of IAD-R1. IAD-R1 employs a progressive two-stage training strategy: First, in the PA-SFT stage, supervised fine-tuning is conducted on pre-trained VLMs using CoT reasoning samples from the Expert-AD dataset to enhance the model's anomaly perception capabilities and establish structured reasoning pathways; Subsequently, in the SC-GRPO stage, reward functions across four dimensions (consistency, accuracy, location, and type) are designed as reinforcement learning objectives to optimize the policy model, thereby improving its anomaly detection and understanding capabilities.}
    \label{framework}
\end{figure*}

\subsection{Expert-AD Dataset}
The field of industrial image anomaly detection lacks high-quality prompt tuning datasets, severely hindering the application of VLMs in this downstream task. Although Anomaly-Instruct-125k proposed by Anomaly-OV~\cite{anomaly-ov} contains a substantial amount of instruction tuning data, the web-crawled data differs significantly from real industrial images in practical scenarios. Moreover, these prompts only include image descriptions and answers, failing to help models develop a complete logic for anomaly inspection. To bridge this gap, we create Expert-AD, a high-quality real industrial scenario tuning dataset containing CoT reasoning. The detailed pipeline generation process is provided in Section A of the Supplementary Material.

The core innovation of Expert-AD lies in the construction of a systematic anomaly detection reasoning framework that implements structured anomaly detection reasoning through observation, comparison, identification, evaluation, and decision-making stages. This reasoning process comprises three core layers: the basic perception layer achieves precise localization of anomaly positions through spatial scanning and key component positioning; the knowledge-driven analysis layer combines industrial standard knowledge to identify anomaly types from dimensions of appearance integrity, surface quality, and structural integrity; the comprehensive decision layer provides final judgments on anomaly existence and evaluates their impact. Based on this reasoning framework, we adopt a two-stage training strategy: the PA-SFT stage trains anomaly perception capabilities through QA pairs containing detailed CoT, establishing mappings between structured reasoning processes and detection results; the SC-GRPO stage enhances the model's analytical and decision-making accuracy in complex scenarios through fine-grained reward function design. In total, our Expert-AD comprises 2.9K QA pairs for PA-SFT stage and 3K QA pairs for SC-GRPO stage.

\subsection{IAD-R1}

% \subsubsection{Overview}
In this paper, we propose IAD-R1, a two-stage post-training framework specifically designed for industrial anomaly detection. 
As shown in Figure \ref{framework}, IAD-R1 comprises two progressive stages: (1) PA-SFT stage, which employs supervised fine-tuning on Expert-AD CoT data to enable the model to acquire structured anomaly analysis thinking patterns and establish effective associations between reasoning processes and detection results; (2) SC-GRPO stage, which guides the model to break through simple pattern memorization, achieving more flexible and accurate anomaly analysis decisions through multi-dimensional reward functions.

\subsubsection{Perception Activation Supervised Fine-Tuning}
In the PA-SFT stage, we activate the model's anomaly perception potential through supervised fine-tuning on the Expert-AD dataset. Training samples adopt a triplet format $(I, p, O)$, where $I$ represents the industrial image, $p$ denotes the text prompt, and $O$ is the conditional output sequence. Through high-quality CoT data training, the model learns to identify anomaly patterns from visual features and generates outputs with different structures based on image content. For normal images, $O=(T,A)$, containing the CoT reasoning process $T$ and final answer $A$; for anomalous images, $O = (T, L, t, A)$, containing the CoT reasoning process $T$, anomaly location $L$, anomaly type $t$, and final answer $A$. This training approach ensures that the model can produce logically coherent structured outputs when processing both normal and anomalous images, avoiding contradictions between the reasoning process and final conclusions. The training objective maximizes the conditional probability of generating the output sequence given the image and prompt:
{\small
\begin{equation}
    \mathcal{L}_{\text{PA-SFT}} = -\mathbb{E}_{(I,p,O) \sim \mathcal{D}_{\text{Expert-AD}}} \sum_{i=1}^{L} \log \pi_\theta(o_i | I, p, o_{<i}),
\end{equation}
}where $D_{Expert-AD}$ is the Expert-AD dataset, $o_i$ is the $i$-th token in the output sequence $O$, $L$ is the sequence length, and $\pi_\theta$ denotes the parameterized model. The trained model serves as the initial policy model for the SC-GRPO stage.

\subsubsection{Structured Control Group Relative Policy Optimization}
We utilize the model $\pi_{PA-SFT}$, obtained through PA-SFT training, as the initial policy model. Targeting the unique characteristics of industrial anomaly detection tasks, we develop an optimized SC-GRPO algorithm. Through carefully designed multi-dimensional reward functions, we achieve consistency between model reasoning and answers while enhancing overall detection accuracy.
\\
\textbf{Multi-dimensional Reward Function Design.}
The core innovation of SC-GRPO lies in the multi-dimensional reward function specifically designed for industrial anomaly detection tasks, comprising four key components.
\\
\textit{Consistency Reward Function.} The consistency reward function $R_{con}$ enhances semantic consistency between model reasoning and detection conclusions by enforcing structured output formats that align with anomaly detection results. We define two standard output patterns corresponding to different anomaly states.
For normal images, the output follows the Normal Pattern $P_{normal}$, containing only the reasoning process and final answer:
{\small
\begin{multline*}
    P_{normal} = \langle think \rangle \text{...} \langle /think \rangle \langle answer \rangle \text{...} \langle /answer \rangle.
\end{multline*}
}For anomalous images, the output follows the Anomalous Pattern $P_{abnormal}$, which includes complete structured information to maintain reasoning consistency:
{\small
\begin{multline*}
    P_{abnormal} = \langle think \rangle \text{...} \langle /think \rangle \langle location \rangle \text{...} \langle /location \rangle\\
    \langle type \rangle \text{...} \langle /type \rangle \langle answer \rangle \text{...} \langle /answer \rangle.
\end{multline*}
}Given model output $o_i$ and corresponding ground truth label $y_i$, the consistency reward function is defined as:
{\small
\begin{equation}
    R_{con}(o_i) = \begin{cases}
1, & \text{if } \text{Match}(o_i, P) \\
0, & \text{otherwise}
\end{cases},
\end{equation}
}where $\text{Match}(o_i, P)$ is a pattern matching function that uses regular expressions to verify whether output $o_i$ strictly conforms to the format requirements of pattern $P$. \\
\textit{Answer Accuracy Reward Function.}
The answer accuracy reward function $R_{acc}$ is used to evaluate the correctness of the model's final anomaly judgment and serves as the core reward signal in the reinforcement learning process. This function directly compares the model's predicted answer and ground truth label.
Given the model's predicted answer $a_{pred}$ and the ground truth answer label $a_{gt}$, the answer accuracy reward function is defined as:
{\small
\begin{equation}
    R_{acc}(a_{pred}, a_{gt}) = \begin{cases} 
1, & \text{if } a_{pred} = a_{gt} \\
0, & \text{otherwise} 
\end{cases}.
\end{equation}
}
This function ensures that rewards are only obtained when the model provides correct anomaly judgments (``Yes" or ``No"), providing a prerequisite condition for subsequent location and type classification rewards. \\
\textit{Location Accuracy Reward Function.}
The location accuracy reward function $R_{loc}$ is used to evaluate the accuracy of the model's anomaly location. This function maps location description text to a standardized $3\times3$ spatial grid we established for comparison, converting textual descriptions into corresponding grid position numbers by parsing location keywords (such as ``left," ``right," ``top," ``bottom," etc.).
Given the model's predicted location description $l_{pred} $ and the ground truth location label $l_{gt}$, the location accuracy reward function is defined as:
{\small
\begin{equation}
    R_{loc}(l_{pred}, l_{gt}) = \begin{cases} 
1, & \text{if } \Phi(l_{pred}) = \Phi(l_{gt}) \\
0, & \text{otherwise} 
\end{cases},
\end{equation}
}where $\Phi(\cdot)$ represents the text-to-grid position mapping function, which provides rewards when the predicted location matches the ground truth location. \\
\textit{Type Accuracy Reward Function.}
The type accuracy reward function $R_{type}$ is used to evaluate the accuracy of the model's anomaly type classification. Considering the diversity and semantic complexity of type descriptions in industrial anomaly detection, using only string matching strategies is overly restrictive and may lead to sparse rewards for the model during the reinforcement learning process, affecting training stability. Therefore, we design a multi-level matching mechanism to improve the stability of reinforcement learning and encourage the model to explore diverse output expressions. Given the matching degree between the predicted type $t_{pred}$ and the ground truth type  $t_{gt}$, the type accuracy reward function is defined as:
{\small
\begin{equation}
    R_{type}(t_{pred}, t_{gt}) = \begin{cases} 
1.0, & \text{if exact match} \\
0.85, & \text{if semantic match} \\
0.6, & \text{if category match} \\
0.4, & \text{if fuzzy match} \\
0.3, & \text{if group match} \\
0, & \text{otherwise} 
\end{cases}.
\end{equation}
}\textbf{Relative Advantage Computation.}
Our SC-GRPO strategy employs intra-group reward normalization to compute relative advantages between responses. This design avoids the training overhead of value function networks in Proximal Policy Optimization~\cite{ppo} while maintaining optimization effectiveness:
{\small
\begin{equation}
    A_i = \frac{R_{\text{SC-GRPO}}(o_i) - \text{mean}\{R_{\text{SC-GRPO}}(o_j)\}_{j=1}^G}{\text{std}\{R_{\text{SC-GRPO}}(o_j)\}_{j=1}^G}.
\end{equation}
}

% 增加Anomaly-R1 Anomaly-OV和IAD-R1 (Qwen3b)后的表格
\begin{table*}[htbp]
\centering
\small
\begin{tabular}{cccccccccc}
\toprule
\multirow{2}{*}{\textbf{Type}} & \multirow{2}{*}{\textbf{Model}} & \multirow{2}{*}{\textbf{Parameter}} & \multicolumn{3}{c}{\textbf{Industrial Workpieces}} & \multicolumn{3}{c}{\textbf{Surface Texture}} & \multirow{2}{*}{\textbf{Average}} \\
\cmidrule(lr){4-6} \cmidrule(lr){7-9}
& & & \textbf{MVTec} & \textbf{MPDD} & \textbf{VisA} & \textbf{DAGM} & \textbf{DTD} & \textbf{SDD} & \\
\midrule
\multirow{6}{*}{Commercial}
& GPT-4o-mini & / & 71.3 & 67.9 & 65.1 & 72.6 & 79.5 & 66.6 & 70.5 \\
& GPT-4o & / & 69.6 & 60.3 & 63.5 & 63.0 & 69.9 & 65.7 & 65.3 \\
& GPT-4.1-nano & / & 74.7 & 61.7 & 60.5 & 62.4 & 78.3 & 50.0 & 64.6 \\
& GPT-4.1-mini & / & 74.0 & 69.8 & 63.4 & 70.7 & 82.1 & 72.1 & 72.0 \\
& GPT-4.1 & / & 81.9 & 66.7 & 69.1 & 81.8 & 90.1 & 79.9 & 78.3 \\
& Claude-Sonnet-4 & / & 67.6 & 65.9 & 63.5 & 69.2 & 88.4 & 81.7 & 72.7 \\
\midrule
\multirow{12}{*}{Open Source}
& LLaVA-OneVision-SI & 0.5B & 50.0 & 50.0 & 50.0 & 50.0 & 54.3 & 50.0 & 50.7 \\
& Anomaly-OV[CVPR 2025] & 0.5B & 50.0 & 50.0 & 50.0 & 50.0 & 53.8 & 50.0 & 50.6 \\
& Qwen2.5-VL-Instruct & 3B & 62.6 & 52.9 & 58.4 & 54.2 & 64.4 & 50.3 & 57.1 \\
& AnomalyR1[arxiv 2025] & 3B & 69.4 & 56.0 & 59.8 & 56.7 & 61.0 & 57.6 & 60.1 \\
& InternVL-2.5 & 4B & 56.6 & 59.1 & 53.7 & 57.7 & 81.3 & 64.1 & 62.1 \\
& Qwen2.5-VL-Instruct & 7B & 66.0 & 56.0 & 58.4 & 57.7 & 59.2 & 67.4 & 60.8 \\
& AnomalyGPT[AAAI 2024] & 7B & 46.6 & 54.2 & 57.3 & 49.6 & 64.1 & 49.5 & 53.6 \\
& LLaVA-OneVision-SI & 7B & \underline{82.0} & 57.0 & 59.6 & 75.4 & 76.8 & 55.1 & 67.7 \\
& Anomaly-OV[CVPR 2025] & 7B & 74.3 & \underline{70.3} & 74.3 & 77.5 & 90.7 & \underline{88.7} & 78.9 \\
& LLaVA-1.5 & 13B & 61.4 & 61.4 & 67.2 & 50.4 & 75.9 & 50.0 & 61.1 \\
& LLaVA-1.6 & 34B & 53.7 & 50.0 & 53.9 & 50.0 & 52.7 & 50.0 & 51.7 \\
& Qwen2.5-VL-Instruct & 72B & 77.4 & 64.7 & 68.2 & 77.9 & 81.9 & 69.0 & 73.2 \\
\midrule
\multirow{4}{*}{IAD-R1}
& IAD-R1(LLaVA-OneVision-SI) & 0.5B & 81.0 & 69.4 & 74.9 & \underline{93.3} & \underline{95.5} & 88.6 & \underline{83.8} \\
& IAD-R1(Qwen2.5-VL-Instruct) & 3B & 77.6 & 59.2 & 69.8 & 85.2 & 89.1 & 83.4 & 77.4 \\
& IAD-R1(Qwen2.5-VL-Instruct) & 7B & 81.9 & 65.8 & \underline{75.4} & 85.2 & 90.8 & 83.4 & 80.4 \\
& IAD-R1(LLaVA-OneVision-SI) & 7B & \textbf{86.7} & \textbf{70.9} & \textbf{78.0} & \textbf{94.8} & \textbf{96.2} & \textbf{90.1} & \textbf{86.1} \\
\bottomrule
\end{tabular}%
\caption{Performance comparison of different models on industrial workpieces and surface texture benchmarks. The best results are highlighted in bold, while the second-best results are underlined.}
\label{model_comparison}
\end{table*}

%实验4_experiment
\section{Experiment}
\subsection{Experiment Setup}
\textbf{Evaluation Details.}
To evaluate the performance of IAD-R1, we select six representative datasets encompassing two major categories: industrial objects (MVTec-AD~\cite{mvtec}, VisA~\cite{visa}, MPDD~\cite{mpdd}) and surface textures (DAGM~\cite{dagm}, DTD~\cite{dtd}, SDD~\cite{sdd}), to simulate complex real-world industrial production scenarios. Considering that overall accuracy is prone to evaluation bias under data imbalance conditions, we adopt balanced accuracy as our evaluation metric. For IAD-R1 models, we extract answers from the structured output tags $\langle answer \rangle \text{...} \langle /answer \rangle$; for baseline models, we directly obtain their output answers for comparison. 
We conduct comprehensive comparisons with both commercial and open-source models, detailed baseline methods are provided in Section C.1 of the Supplementary Material. \\
\textbf{Implementation Details.}
To validate the generalizability of IAD-R1, we employ 7 VLMs with different architectures and parameter scales as backbone models, including Qwen2-VL-2B~\cite{qwen2-vl}, Qwen2.5-VL (3B, 7B)~\cite{qwen2.5-vl}, LLaVA-1.5-7B~\cite{llava-1.5}, LLaVA-OneVision-SI (0.5B, 7B)~\cite{llava-ov-si}, and LLaVA-1.6-8B~\cite{llava-next}. All experiments are conducted on 4$\times$A100 GPUs. More Implementation details can be found in Section C.1 of the Supplementary Material.
\subsection{Main Results}
\textbf{Model Comparison.} As shown in Table \ref{model_comparison}, IAD-R1(LLaVA-OneVision-SI-7B) achieved the highest average accuracy of 86.1\%, representing improvements of 7.2\% and 7.8\% over the best open-source model Anomaly-OV~\cite{anomaly-ov} and commercial model GPT-4.1~\cite{gpt4.1}, respectively. Notably, small-parameter models trained with IAD-R1 demonstrated exceptional performance: the IAD-R1(Qwen2.5-VL-Instruct-3B) model surpassed the 72B model of the same series by 4.2\%, while the IAD-R1(LLaVA-OneVision-SI-0.5B) even outperformed all open-source and commercial models. These results indicate that our IAD-R1 post-training strategy is more effective than simply increasing parameters, providing a parameter-efficient solution for industrial applications in resource-constrained environments. Detailed comparisons can be found in Section C.5 of the Supplementary Material.

\subsection{Ablation Results}
\textbf{Ablation on Data Activation Method.}
To verify the role of CoT responses on model performance, we compare different data preparation methods during the PA-SFT stage. As shown in Table \ref{ablation_sft}, ``Expert-AD" refers to fine-tuning with our Expert-AD dataset containing complete structured reasoning processes, and ``Original" refers to fine-tuning with the same images but using direct answers without reasoning processes. Using Expert-AD data in the PA-SFT stage significantly outperforms original data, with this result being validated across all three models. Notably, fine-tuning with original data even exhibits performance degradation in some cases, indicating that simple question-answer training may lead to inadequate learning or overfitting, while the CoT responses in Expert-AD can guide models to learn deeper anomaly analysis logic. Additional experimental results can be found in Section C.3 of the Supplementary Material.

\begin{table}[htbp]
\centering
\small
\setlength{\tabcolsep}{1.5mm}
\begin{tabular}{ccccc}
\toprule
\textbf{Model} & \textbf{Data} & \textbf{0-shot} & \textbf{1-shot} & \textbf{Average} \\
\midrule
\multirow{3}{*}{\begin{tabular}{@{}c@{}}LLaVA-OneVision-SI\\(0.5B)\end{tabular}}
& Base & 50.7 & 50.0 & 50.4 \\
& Original & 50.9 & 50.0 & 50.5 \\
& Expert-AD & \textbf{82.7} & \textbf{69.5} & \textbf{76.1} \\
\multirow{3}{*}{\begin{tabular}{@{}c@{}}Qwen2.5-VL-Instruct\\(3B)\end{tabular}}
& Base & 57.1 & 59.9 & 58.5 \\
& Original & 65.8 & 59.5 & 62.7 \\
& Expert-AD & \textbf{73.1} & \textbf{73.5} & \textbf{73.3} \\
\multirow{3}{*}{\begin{tabular}{@{}c@{}}LLaVA-OneVision-SI\\(7B)\end{tabular}}
& Base & 67.7 & 60.6 & 64.2 \\
& Original & 68.4 & 62.3 & 65.4 \\
& Expert-AD & \textbf{85.9} & \textbf{74.9} & \textbf{80.4} \\
\bottomrule
\end{tabular}%
\caption{Results of different data preparation in PA-SFT.}
\label{ablation_sft}
\end{table}
\noindent
\textbf{Ablation on Reward Strategy.}
To verify the importance of multi-dimensional reward functions, we compare SC-GRPO and the original reward strategy on three models fine-tuned with PA-SFT, where the original reward strategy employs only the correctness of final answers as the reward signal. As shown in Table \ref{ablation_grpo}, experimental results demonstrate that SC-GRPO significantly outperforms the original reward strategy across all models. The original reward strategy exhibits performance degradation primarily because its single reward design easily leads to sparse reward problems and cannot distinguish between ``correct answers with clear reasoning" and ``correct answers with chaotic reasoning". In contrast, SC-GRPO provides more refined optimization guidance through multi-dimensional reward mechanisms, achieving more stable reinforcement learning optimization.

Table \ref{ablation_reward} presents an ablation study examining the contribution of each reward function in SC-GRPO. Each individual reward function makes a meaningful contribution to the overall performance. Furthermore, the integration of all reward functions achieves optimal performance across different datasets. Additional experimental results can be found in Section C.3 of the Supplementary Material.

\begin{table}[ht]
\centering
\small
\setlength{\tabcolsep}{1.5mm}
\begin{tabular}{ccccc}
\toprule
\textbf{Model} & \textbf{Strategy} & \textbf{0-shot} & \textbf{1-shot} & \textbf{Average} \\
\midrule
\multirow{3}{*}{\begin{tabular}{@{}c@{}}LLaVA-OneVision-SI\\(0.5B)\end{tabular}}
& Fine-tuned & 82.7 & \textbf{69.5} & 76.1 \\
& Original & 78.1 & 59.0 & 68.6 \\
& SC-GRPO & \textbf{83.8} & 68.8 & \textbf{76.3} \\
\multirow{3}{*}{\begin{tabular}{@{}c@{}}Qwen2.5-VL-Instruct\\(3B)\end{tabular}}
& Fine-tuned & 73.1 & 73.5 & 73.3 \\
& Original & 70.0 & 71.9 & 71.0 \\
& SC-GRPO & \textbf{77.4} & \textbf{76.4} & \textbf{76.9} \\
\multirow{3}{*}{\begin{tabular}{@{}c@{}}LLaVA-OneVision-SI\\(7B)\end{tabular}}
& Fine-tuned & 85.9 & 74.9 & 80.4 \\
& Original & 78.5 & 65.8 & 72.2 \\
& SC-GRPO & \textbf{86.1} & \textbf{75.8} & \textbf{81.0} \\
\bottomrule
\end{tabular}
\caption{Results of different reward strategy in SC-GRPO.}
\label{ablation_grpo}
\end{table}
\noindent

\begin{figure}[!htpb]
    \centering
    \includegraphics[width=1\linewidth]{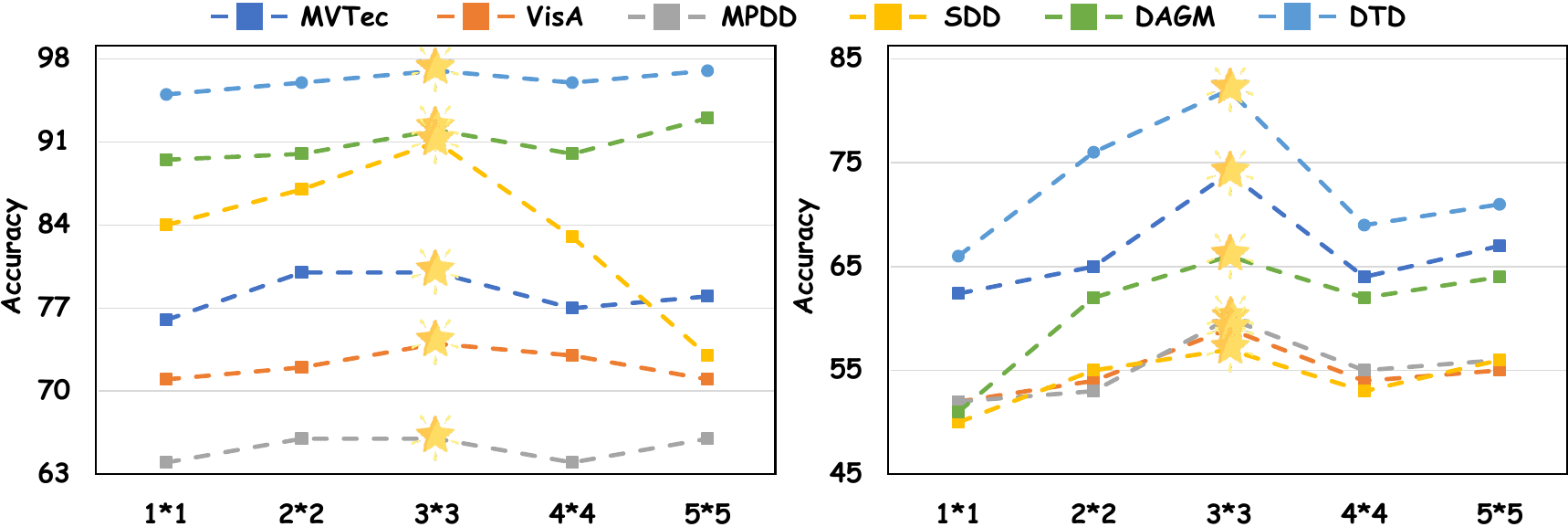}
    \caption{Ablation on grid size.}
    \label{grid}
\end{figure}

\textbf{Discussion about Grid Size.}
Figure \ref{grid} demonstrates the impact of grid partition size in the location accuracy reward function on model performance (left: LLaVA-OneVision-SI-0.5B;right: Qwen2-VL-Instruct-2B). The 3$\times$3 grid partition achieves optimal performance, providing sufficient anomaly localization details while avoiding noise interference from overly fine-grained partitioning.

\begin{table*}[htbp]
\centering
\small
\setlength{\tabcolsep}{1.5mm}
\begin{tabular}{cccccccccccc}
\toprule
\multirow{3}{*}{\textbf{Model}} & \multirow{3}{*}{\textbf{Parameter}} & \multicolumn{2}{c}{\textbf{Strategy}} & \multicolumn{6}{c}{\textbf{0-shot}} & \multicolumn{2}{c}{\textbf{1-shot}} \\
\cmidrule(lr){3-4} \cmidrule(lr){5-10} \cmidrule(lr){11-12}
& & \textbf{PA-SFT} & \textbf{SC-GRPO} & \textbf{MVTec} & \textbf{DAGM} & \textbf{DTD} & \textbf{SDD} & \textbf{MPDD} & \textbf{VisA} & \textbf{MVTec} & \textbf{VisA} \\
\midrule
\multirow{3}{*}{LLaVA-OneVision-SI} 
& \multirow{3}{*}{0.5B} & & & 50.0 & 50.0 & 54.3 & 50.0 & 50.0 & 50.0 & 50.0 & 49.9 \\
& & \checkmark & & 79.4 & 91.3 & \textbf{96.0} & \textbf{90.7} & 65.3 & 73.4 & \textbf{72.6} & 66.4 \\
& & \checkmark & \checkmark & \textbf{81.0} & \textbf{93.3} & 95.5 & 88.6 & \textbf{69.4} & \textbf{74.9} & 70.8 & \textbf{66.9} \\
\multirow{3}{*}{Qwen2-VL-Instruct} 
& \multirow{3}{*}{2B} & & & 63.5 & 54.3 & 59.3 & 57.2 & 55.0 & 59.6 & 69.9 & 59.6 \\
& & \checkmark & & 73.5 & 65.4 & 81.6 & 56.3 & 59.3 & 58.8 & 77.0 & 60.4 \\
& & \checkmark & \checkmark & \textbf{77.3} & \textbf{73.8} & \textbf{84.0} & \textbf{62.8} & \textbf{69.5} & \textbf{67.7} & \textbf{78.5} & \textbf{69.7} \\
\multirow{6}{*}{Qwen2.5-VL-Instruct} 
& \multirow{3}{*}{3B} & & & 62.6 & 54.2 & 64.4 & 50.3 & 52.9 & 58.4 & 60.1 & 59.6 \\
& & \checkmark & & 73.3 & 82.8 & 87.7 & 76.3 & 53.4 & 65.1 & 75.9 & 71.0 \\
& & \checkmark & \checkmark & \textbf{77.6} & \textbf{85.2} & \textbf{89.1} & \textbf{83.4} & \textbf{59.2} & \textbf{69.8} & \textbf{78.6} & \textbf{74.1} \\
& \multirow{3}{*}{7B} & & & 66.0 & 57.7 & 59.2 & 67.4 & 56.0 & 58.4 & 81.0 & 69.8 \\
& & \checkmark & & 78.2 & 83.8 & 88.0 & 80.4 & 63.9 & 70.4 & 82.6 & 70.5 \\
& & \checkmark & \checkmark & \textbf{81.9} & \textbf{85.2} & \textbf{90.8} & \textbf{83.4} & \textbf{65.8} & \textbf{75.4} & \textbf{85.5} & \textbf{77.7} \\
\multirow{3}{*}{LLaVA-1.5} 
& \multirow{3}{*}{7B} & & & 67.4 & 63.7 & 79.3 & 58.0 & 62.0 & 71.1 & 59.8 & 53.6 \\
& & \checkmark & & 78.1 & 69.6 & 84.0 & \textbf{83.9} & 60.9 & 69.8 & 76.1 & 67.2 \\
& & \checkmark & \checkmark & \textbf{79.4} & \textbf{75.4} & \textbf{86.7} & 83.6 & \textbf{62.3} & \textbf{72.9} & \textbf{77.8} & \textbf{71.8} \\
\multirow{3}{*}{LLaVA-OneVision-SI} 
& \multirow{3}{*}{7B} & & & 82.0 & 75.4 & 76.8 & 55.1 & 57.0 & 59.6 & 67.5 & 53.7 \\
& & \checkmark & & \textbf{86.7} & 94.3 & 96.1 & 89.5 & 70.5 & \textbf{78.0} & 79.3 & 70.4 \\
& & \checkmark & \checkmark & \textbf{86.7} & \textbf{94.8} & \textbf{96.2} & \textbf{90.1} & \textbf{70.9} & \textbf{78.0} & \textbf{80.0} & \textbf{71.5} \\
\multirow{3}{*}{LLaVA-1.6} 
& \multirow{3}{*}{8B} & & & 62.7 & 52.0 & 73.6 & 50.0 & \textbf{66.2} & 56.7 & 51.5 & 52.6 \\
& & \checkmark & & 83.1 & 73.4 & 80.0 & 69.4 & \textbf{66.2} & 68.2 & 80.1 & 63.7 \\
& & \checkmark & \checkmark & \textbf{84.7} & \textbf{79.0} & \textbf{88.0} & \textbf{84.5} & 64.1 & \textbf{70.4} & \textbf{78.0} & \textbf{64.4} \\
\bottomrule
\end{tabular}%
\caption{Performance gains from using IAD-R1 on VLMs with different architectures and parameter scales.}
\label{model_promotion}
\end{table*}
\noindent
\textbf{Model Promotion.} Table \ref{model_promotion} provides a detailed presentation of the gains brought by IAD-R1 relative to the backbone across different test settings and datasets. The PA-SFT stage significantly activates the anomaly perception capabilities, while the SC-GRPO stage further optimizes output quality. Across the parameter range from 0.5B to 8B, all tested models in the Qwen and LLaVA series achieved significant performance improvements from the two-stage training, validating the broad applicability of IAD-R1 as a general industrial anomaly detection post-training framework. Notably, IAD-R1 demonstrates more pronounced improvement effects in 0-shot scenarios, which aligns with our training design using single-image data.

\begin{table}[!ht]
\centering
\small
\setlength{\tabcolsep}{0.8mm}
\begin{tabular}{ccccccccc}
\toprule
\textbf{$R_{con}$} & \textbf{$R_{loc}$} & \textbf{$R_{type}$} & \textbf{MVTec} & \textbf{VisA} & \textbf{DAGM} & \textbf{MPDD} & \textbf{SDD} & \textbf{DTD} \\
\midrule
 &  &  & 72.7 & 59.0 & 72.5 & 55.5 & 79.3 & 81.0 \\
 & $\checkmark$ &  & 76.3 & 66.3 & 83.9 & 57.9 & 77.4 & 88.4 \\
 &  & $\checkmark$ & 77.4 & 67.0 & 84.8 & 57.9 & 81.6 & 88.1 \\
$\checkmark$ & $\checkmark$ &  & 75.4 & 67.0 & \textbf{85.8} & 58.6 & 79.5 & 90.0 \\
$\checkmark$ &  & $\checkmark$ & 76.7 & 67.5 & 85.1 & 56.5 & 80.8 & \textbf{89.1} \\
$\checkmark$ & $\checkmark$ & $\checkmark$ & \textbf{77.6} & \textbf{69.8} & 85.2 & \textbf{59.2} & \textbf{83.4} & \textbf{89.1} \\
\bottomrule
\end{tabular}%
\caption{Ablation of reward function contributions.}
\label{ablation_reward}
\end{table}

\subsection{Analysis of Reasoning and Answer Consistency}
\begin{figure}[!ht]
    \centering
    \includegraphics[width=1\linewidth]{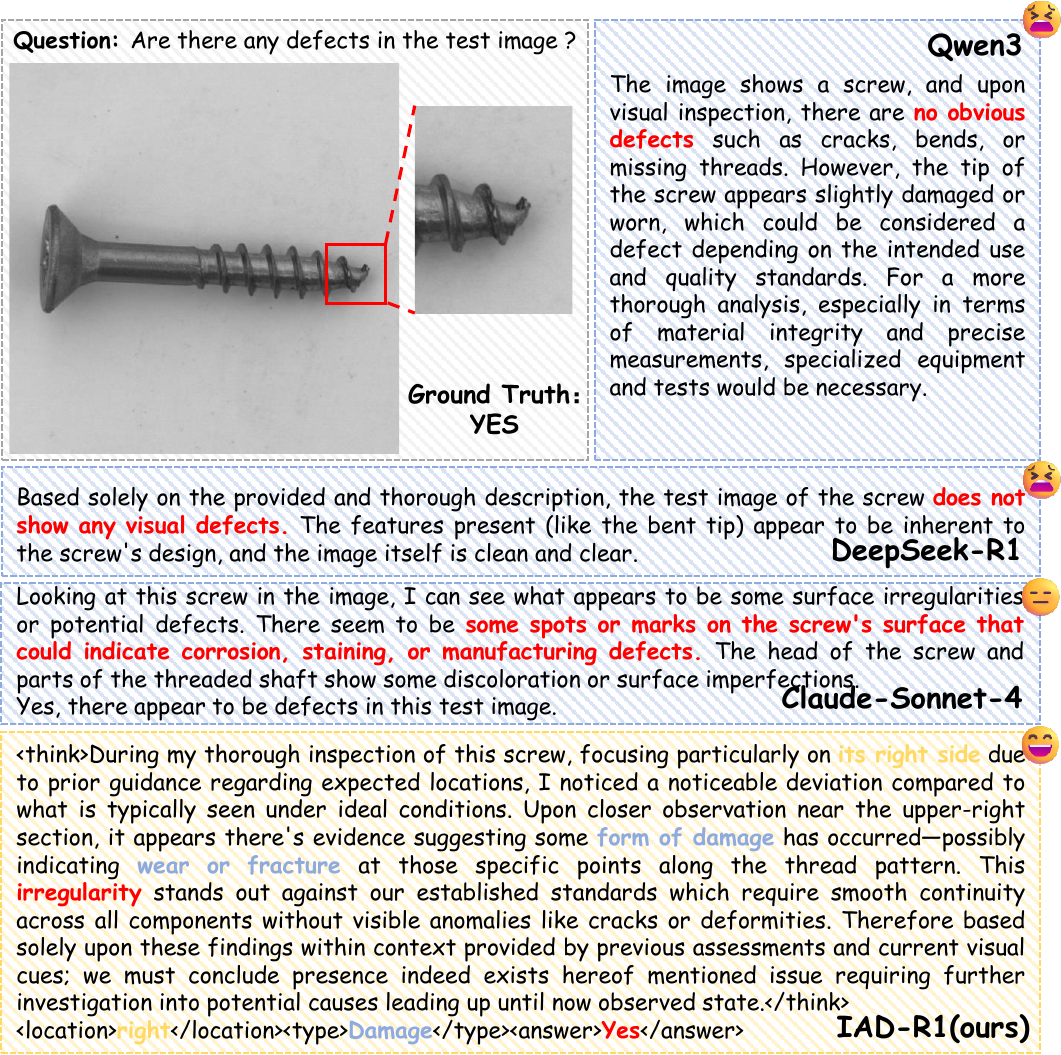}
    \caption{Comparison of model output.}
    \label{visualization}
\end{figure}

Figure \ref{visualization} illustrates the performance of different models on an anomaly detection task. Qwen3~\cite{qwen3} and DeepSeek-R1~\cite{deepseek-r1} provide incorrect answers, while Claude-Sonnet-4~\cite{claude4}, although producing the correct answer, exhibits flawed reasoning processes. In contrast, IAD-R1 not only delivers the correct answer but also ensures consistency between the reasoning process and the final answer. Additional model outputs can be found in Section C.6 of the Supplementary Material.

% 结论部分需要修改，提出什么方法/架构，能解决什么问题（P6）

\section{Conclusion}
In this paper, we propose IAD-R1 to address key challenges faced by VLMs in industrial anomaly detection. We construct the Expert-AD dataset, contributing the first high-quality CoT reasoning data resource for this domain. Additionally, we design PA-SFT and SC-GRPO methods that enable the model to transcend traditional pattern matching approaches through structured reasoning training, achieving a leap from ``Anomaly Perception" to ``Anomaly Interpretation". Experimental results demonstrate that IAD-R1 achieves significant performance improvements over baseline models across all benchmark datasets, providing a viable and effective technical pathway for advancing VLM applications in industrial anomaly detection scenarios.


\bibliography{aaai2026}

\begin{thebibliography}{56}
\providecommand{\natexlab}[1]{#1}

\bibitem[{Anthropic(2025)}]{claude4}
Anthropic. 2025.
\newblock Introducing Claude 4.
\newblock \url{https://www.anthropic.com/news/claude-4}.
\newblock Accessed: 2025-07-28.

\bibitem[{Aota, Tong, and Okatani(2023)}]{dtd}
Aota, T.; Tong, L. T.~T.; and Okatani, T. 2023.
\newblock Zero-shot versus many-shot: Unsupervised texture anomaly detection.
\newblock In \emph{Proceedings of the IEEE/CVF Winter Conference on Applications of Computer Vision}, 5564--5572.

\bibitem[{Bae, Lee, and Kim(2023)}]{pni}
Bae, J.; Lee, J.-H.; and Kim, S. 2023.
\newblock Pni: industrial anomaly detection using position and neighborhood information.
\newblock In \emph{Proceedings of the IEEE/CVF International Conference on Computer Vision}, 6373--6383.

\bibitem[{Bai et~al.(2025)Bai, Chen, Liu, Wang, Ge, Song, Dang, Wang, Wang, Tang et~al.}]{qwen2.5-vl}
Bai, S.; Chen, K.; Liu, X.; Wang, J.; Ge, W.; Song, S.; Dang, K.; Wang, P.; Wang, S.; Tang, J.; et~al. 2025.
\newblock Qwen2. 5-vl technical report.
\newblock \emph{arXiv preprint arXiv:2502.13923}.

\bibitem[{Bergmann et~al.(2019)Bergmann, Fauser, Sattlegger, and Steger}]{mvtec}
Bergmann, P.; Fauser, M.; Sattlegger, D.; and Steger, C. 2019.
\newblock MVTec AD--A comprehensive real-world dataset for unsupervised anomaly detection.
\newblock In \emph{Proceedings of the IEEE/CVF conference on computer vision and pattern recognition}, 9592--9600.

\bibitem[{Cao et~al.(2024)Cao, Zhang, Frittoli, Cheng, Shen, and Boracchi}]{adaclip}
Cao, Y.; Zhang, J.; Frittoli, L.; Cheng, Y.; Shen, W.; and Boracchi, G. 2024.
\newblock Adaclip: Adapting clip with hybrid learnable prompts for zero-shot anomaly detection.
\newblock In \emph{European Conference on Computer Vision}, 55--72. Springer.

\bibitem[{Chao et~al.(2025)Chao, Liu, Tang, and Wu}]{anomalyr1}
Chao, Y.; Liu, J.; Tang, J.; and Wu, G. 2025.
\newblock Anomalyr1: A grpo-based end-to-end mllm for industrial anomaly detection.
\newblock \emph{arXiv preprint arXiv:2504.11914}.

\bibitem[{Chen et~al.(2025)Chen, Chen, Imani, and Imani}]{echo}
Chen, Z.; Chen, H.; Imani, M.; and Imani, F. 2025.
\newblock Can multimodal large language models be guided to improve industrial anomaly detection?
\newblock \emph{arXiv preprint arXiv:2501.15795}.

\bibitem[{Chen et~al.(2024)Chen, Wang, Cao, Liu, Gao, Cui, Zhu, Ye, Tian, Liu et~al.}]{internvl2.5}
Chen, Z.; Wang, W.; Cao, Y.; Liu, Y.; Gao, Z.; Cui, E.; Zhu, J.; Ye, S.; Tian, H.; Liu, Z.; et~al. 2024.
\newblock Expanding performance boundaries of open-source multimodal models with model, data, and test-time scaling.
\newblock \emph{arXiv preprint arXiv:2412.05271}.

\bibitem[{Deng et~al.(2023)Deng, Zhang, Bao, and Li}]{anovl}
Deng, H.; Zhang, Z.; Bao, J.; and Li, X. 2023.
\newblock Anovl: Adapting vision-language models for unified zero-shot anomaly localization.
\newblock \emph{arXiv preprint arXiv:2308.15939}, 2(5).

\bibitem[{Gu et~al.(2024{\natexlab{a}})Gu, Zhu, Zhu, Chen, Li, Tang, and Wang}]{filo}
Gu, Z.; Zhu, B.; Zhu, G.; Chen, Y.; Li, H.; Tang, M.; and Wang, J. 2024{\natexlab{a}}.
\newblock Filo: Zero-shot anomaly detection by fine-grained description and high-quality localization.
\newblock In \emph{Proceedings of the 32nd ACM International Conference on Multimedia}, 2041--2049.

\bibitem[{Gu et~al.(2024{\natexlab{b}})Gu, Zhu, Zhu, Chen, Tang, and Wang}]{anomalygpt}
Gu, Z.; Zhu, B.; Zhu, G.; Chen, Y.; Tang, M.; and Wang, J. 2024{\natexlab{b}}.
\newblock Anomalygpt: Detecting industrial anomalies using large vision-language models.
\newblock In \emph{Proceedings of the AAAI conference on artificial intelligence}, volume~38, 1932--1940.

\bibitem[{Guo et~al.(2025)Guo, Yang, Zhang, Song, Zhang, Xu, Zhu, Ma, Wang, Bi et~al.}]{deepseek-r1}
Guo, D.; Yang, D.; Zhang, H.; Song, J.; Zhang, R.; Xu, R.; Zhu, Q.; Ma, S.; Wang, P.; Bi, X.; et~al. 2025.
\newblock Deepseek-r1: Incentivizing reasoning capability in llms via reinforcement learning.
\newblock \emph{arXiv preprint arXiv:2501.12948}.

\bibitem[{He et~al.(2024)He, Zhang, Chen, Chen, Li, Chen, Wang, Wang, and Xie}]{diad}
He, H.; Zhang, J.; Chen, H.; Chen, X.; Li, Z.; Chen, X.; Wang, Y.; Wang, C.; and Xie, L. 2024.
\newblock A diffusion-based framework for multi-class anomaly detection.
\newblock In \emph{Proceedings of the AAAI conference on artificial intelligence}, volume~38, 8472--8480.

\bibitem[{Hu et~al.(2022)Hu, Shen, Wallis, Allen-Zhu, Li, Wang, Wang, Chen et~al.}]{lora}
Hu, E.~J.; Shen, Y.; Wallis, P.; Allen-Zhu, Z.; Li, Y.; Wang, S.; Wang, L.; Chen, W.; et~al. 2022.
\newblock Lora: Low-rank adaptation of large language models.
\newblock \emph{ICLR}, 1(2): 3.

\bibitem[{Hurst et~al.(2024)Hurst, Lerer, Goucher, Perelman, Ramesh, Clark, Ostrow, Welihinda, Hayes, Radford et~al.}]{gpt-4o}
Hurst, A.; Lerer, A.; Goucher, A.~P.; Perelman, A.; Ramesh, A.; Clark, A.; Ostrow, A.; Welihinda, A.; Hayes, A.; Radford, A.; et~al. 2024.
\newblock Gpt-4o system card.
\newblock \emph{arXiv preprint arXiv:2410.21276}.

\bibitem[{Jeong et~al.(2023)Jeong, Zou, Kim, Zhang, Ravichandran, and Dabeer}]{winclip}
Jeong, J.; Zou, Y.; Kim, T.; Zhang, D.; Ravichandran, A.; and Dabeer, O. 2023.
\newblock Winclip: Zero-/few-shot anomaly classification and segmentation.
\newblock In \emph{Proceedings of the IEEE/CVF Conference on Computer Vision and Pattern Recognition}, 19606--19616.

\bibitem[{Jia et~al.(2022)Jia, Tang, Chen, Cardie, Belongie, Hariharan, and Lim}]{mpdd}
Jia, M.; Tang, L.; Chen, B.-C.; Cardie, C.; Belongie, S.; Hariharan, B.; and Lim, S.-N. 2022.
\newblock Visual prompt tuning.
\newblock In \emph{European conference on computer vision}, 709--727. Springer.

\bibitem[{Jiang et~al.(2024)Jiang, Li, Deng, Liu, Gao, Zhou, Li, Wang, and Zheng}]{mmad}
Jiang, X.; Li, J.; Deng, H.; Liu, Y.; Gao, B.-B.; Zhou, Y.; Li, J.; Wang, C.; and Zheng, F. 2024.
\newblock Mmad: A comprehensive benchmark for multimodal large language models in industrial anomaly detection.
\newblock \emph{arXiv preprint arXiv:2410.09453}.

\bibitem[{Jin et~al.(2025)Jin, Feng, Mou, Lakemeyer, Decker, Simons, and Stegmaier}]{logicad}
Jin, E.; Feng, Q.; Mou, Y.; Lakemeyer, G.; Decker, S.; Simons, O.; and Stegmaier, J. 2025.
\newblock Logicad: Explainable anomaly detection via vlm-based text feature extraction.
\newblock In \emph{Proceedings of the AAAI Conference on Artificial Intelligence}, volume~39, 4129--4137.

\bibitem[{Kwon et~al.(2025)Kwon, Moon, Oh, and Yoon}]{logicqa}
Kwon, Y.; Moon, D.; Oh, Y.; and Yoon, H. 2025.
\newblock LogicQA: Logical Anomaly Detection with Vision Language Model Generated Questions.
\newblock \emph{arXiv preprint arXiv:2503.20252}.

\bibitem[{Li et~al.(2024{\natexlab{a}})Li, Zhang, Guo, Zhang, Li, Zhang, Zhang, Zhang, Li, Liu et~al.}]{llava-ov-si}
Li, B.; Zhang, Y.; Guo, D.; Zhang, R.; Li, F.; Zhang, H.; Zhang, K.; Zhang, P.; Li, Y.; Liu, Z.; et~al. 2024{\natexlab{a}}.
\newblock Llava-onevision: Easy visual task transfer.
\newblock \emph{arXiv preprint arXiv:2408.03326}.

\bibitem[{Li et~al.(2024{\natexlab{b}})Li, Zhang, Zhang, Zhang, Li, Li, Ma, and Li}]{llava-next}
Li, F.; Zhang, R.; Zhang, H.; Zhang, Y.; Li, B.; Li, W.; Ma, Z.; and Li, C. 2024{\natexlab{b}}.
\newblock Llava-next-interleave: Tackling multi-image, video, and 3d in large multimodal models.
\newblock \emph{arXiv preprint arXiv:2407.07895}.

\bibitem[{Li et~al.(2025{\natexlab{a}})Li, Chu, Chen, Xie, Shan, and Zhao}]{lad-reasoner}
Li, W.; Chu, G.; Chen, J.; Xie, G.-S.; Shan, C.; and Zhao, F. 2025{\natexlab{a}}.
\newblock Lad-reasoner: Tiny multimodal models are good reasoners for logical anomaly detection.
\newblock \emph{arXiv preprint arXiv:2504.12749}.

\bibitem[{Li et~al.(2024{\natexlab{c}})Li, Zhang, Tan, Chen, Qu, Xie, and Ma}]{promptad}
Li, X.; Zhang, Z.; Tan, X.; Chen, C.; Qu, Y.; Xie, Y.; and Ma, L. 2024{\natexlab{c}}.
\newblock Promptad: Learning prompts with only normal samples for few-shot anomaly detection.
\newblock In \emph{Proceedings of the IEEE/CVF Conference on Computer Vision and Pattern Recognition}, 16838--16848.

\bibitem[{Li et~al.(2024{\natexlab{d}})Li, Feng, Chen, Chen, Wang, Hu, Qu, Zhou et~al.}]{vpdm}
Li, Y.; Feng, Y.; Chen, B.; Chen, W.; Wang, Y.; Hu, X.; Qu, C.; Zhou, M.; et~al. 2024{\natexlab{d}}.
\newblock Vague prototype-oriented diffusion model for multi-class anomaly detection.
\newblock In \emph{Forty-first International Conference on Machine Learning}.

\bibitem[{Li et~al.(2023)Li, Wang, Yuan, Liu, Zhao, Guo, Xu, Shi, and Zuo}]{myriad}
Li, Y.; Wang, H.; Yuan, S.; Liu, M.; Zhao, D.; Guo, Y.; Xu, C.; Shi, G.; and Zuo, W. 2023.
\newblock Myriad: Large multimodal model by applying vision experts for industrial anomaly detection.
\newblock \emph{arXiv preprint arXiv:2310.19070}.

\bibitem[{Li et~al.(2025{\natexlab{b}})Li, Yuan, Wang, Li, Liu, Xu, Shi, and Zuo}]{triad}
Li, Y.; Yuan, S.; Wang, H.; Li, Q.; Liu, M.; Xu, C.; Shi, G.; and Zuo, W. 2025{\natexlab{b}}.
\newblock Triad: Empowering lmm-based anomaly detection with vision expert-guided visual tokenizer and manufacturing process.
\newblock \emph{arXiv preprint arXiv:2503.13184}.

\bibitem[{Li et~al.(2024{\natexlab{e}})Li, Zhang, Li, and Lao}]{one-to-normal}
Li, Y.; Zhang, S.; Li, K.; and Lao, Q. 2024{\natexlab{e}}.
\newblock One-to-normal: Anomaly personalization for few-shot anomaly detection.
\newblock \emph{Advances in Neural Information Processing Systems}, 37: 78371--78393.

\bibitem[{Liu et~al.(2024{\natexlab{a}})Liu, Li, Li, and Lee}]{llava-1.5}
Liu, H.; Li, C.; Li, Y.; and Lee, Y.~J. 2024{\natexlab{a}}.
\newblock Improved baselines with visual instruction tuning.
\newblock In \emph{Proceedings of the IEEE/CVF conference on computer vision and pattern recognition}, 26296--26306.

\bibitem[{Liu et~al.(2024{\natexlab{b}})Liu, Xie, Wang, Li, Wang, Zheng, and Jin}]{iad-survey}
Liu, J.; Xie, G.; Wang, J.; Li, S.; Wang, C.; Zheng, F.; and Jin, Y. 2024{\natexlab{b}}.
\newblock Deep industrial image anomaly detection: A survey.
\newblock \emph{Machine Intelligence Research}, 21(1): 104--135.

\bibitem[{Ma et~al.(2025)Ma, Zhang, Yao, Tang, Wu, Li, Yan, Jiang, and Zhou}]{aaclip}
Ma, W.; Zhang, X.; Yao, Q.; Tang, F.; Wu, C.; Li, Y.; Yan, R.; Jiang, Z.; and Zhou, S.~K. 2025.
\newblock Aa-clip: Enhancing zero-shot anomaly detection via anomaly-aware clip.
\newblock In \emph{Proceedings of the Computer Vision and Pattern Recognition Conference}, 4744--4754.

\bibitem[{OpenAI(2025)}]{gpt4.1}
OpenAI. 2025.
\newblock Introducing GPT-4.1 in the API.
\newblock \url{https://openai.com/index/gpt-4-1/}.
\newblock Accessed: 2025-07-28.

\bibitem[{Qu et~al.(2025)Qu, Tao, Gong, Qu, Chen, Zhang, Wang, and Ding}]{bayesian}
Qu, Z.; Tao, X.; Gong, X.; Qu, S.; Chen, Q.; Zhang, Z.; Wang, X.; and Ding, G. 2025.
\newblock Bayesian Prompt Flow Learning for Zero-Shot Anomaly Detection.
\newblock In \emph{Proceedings of the Computer Vision and Pattern Recognition Conference}, 30398--30408.

\bibitem[{Qu et~al.(2024)Qu, Tao, Prasad, Shen, Zhang, Gong, and Ding}]{vcp-clip}
Qu, Z.; Tao, X.; Prasad, M.; Shen, F.; Zhang, Z.; Gong, X.; and Ding, G. 2024.
\newblock Vcp-clip: A visual context prompting model for zero-shot anomaly segmentation.
\newblock In \emph{European Conference on Computer Vision}, 301--317. Springer.

\bibitem[{Radford et~al.(2021)Radford, Kim, Hallacy, Ramesh, Goh, Agarwal, Sastry, Askell, Mishkin, Clark et~al.}]{clip}
Radford, A.; Kim, J.~W.; Hallacy, C.; Ramesh, A.; Goh, G.; Agarwal, S.; Sastry, G.; Askell, A.; Mishkin, P.; Clark, J.; et~al. 2021.
\newblock Learning transferable visual models from natural language supervision.
\newblock In \emph{International conference on machine learning}, 8748--8763. PmLR.

\bibitem[{Roth et~al.(2022)Roth, Pemula, Zepeda, Sch{\"o}lkopf, Brox, and Gehler}]{patchcore}
Roth, K.; Pemula, L.; Zepeda, J.; Sch{\"o}lkopf, B.; Brox, T.; and Gehler, P. 2022.
\newblock Towards total recall in industrial anomaly detection.
\newblock In \emph{Proceedings of the IEEE/CVF conference on computer vision and pattern recognition}, 14318--14328.

\bibitem[{Schulman et~al.(2017)Schulman, Wolski, Dhariwal, Radford, and Klimov}]{ppo}
Schulman, J.; Wolski, F.; Dhariwal, P.; Radford, A.; and Klimov, O. 2017.
\newblock Proximal policy optimization algorithms.
\newblock \emph{arXiv preprint arXiv:1707.06347}.

\bibitem[{Shao et~al.(2024)Shao, Wang, Zhu, Xu, Song, Bi, Zhang, Zhang, Li, Wu et~al.}]{grpo}
Shao, Z.; Wang, P.; Zhu, Q.; Xu, R.; Song, J.; Bi, X.; Zhang, H.; Zhang, M.; Li, Y.; Wu, Y.; et~al. 2024.
\newblock Deepseekmath: Pushing the limits of mathematical reasoning in open language models, 2024.
\newblock \emph{URL https://arxiv. org/abs/2402.03300}, 2(3): 5.

\bibitem[{Shen et~al.(2025)Shen, Liu, Li, Fang, Ma, Liao, Shen, Zhang, Zhao, Zhang et~al.}]{vlm-r1}
Shen, H.; Liu, P.; Li, J.; Fang, C.; Ma, Y.; Liao, J.; Shen, Q.; Zhang, Z.; Zhao, K.; Zhang, Q.; et~al. 2025.
\newblock Vlm-r1: A stable and generalizable r1-style large vision-language model.
\newblock \emph{arXiv preprint arXiv:2504.07615}.

\bibitem[{Tabernik et~al.(2020)Tabernik, {\v{S}}ela, Skvar{\v{c}}, and Sko{\v{c}}aj}]{sdd}
Tabernik, D.; {\v{S}}ela, S.; Skvar{\v{c}}, J.; and Sko{\v{c}}aj, D. 2020.
\newblock Segmentation-based deep-learning approach for surface-defect detection.
\newblock \emph{Journal of Intelligent Manufacturing}, 31(3): 759--776.

\bibitem[{Team(2024)}]{qwen-vl-max}
Team, Q. 2024.
\newblock Introducing Qwen-VL.
\newblock https://qwenlm.github.io/blog/qwen-vl/.
\newblock Accessed: 2025-07-28.

\bibitem[{Wang et~al.(2024{\natexlab{a}})Wang, Zhu, Gao, Gan, Zhang, Gu, Qian, Chen, and Ma}]{real-iad}
Wang, C.; Zhu, W.; Gao, B.-B.; Gan, Z.; Zhang, J.; Gu, Z.; Qian, S.; Chen, M.; and Ma, L. 2024{\natexlab{a}}.
\newblock Real-iad: A real-world multi-view dataset for benchmarking versatile industrial anomaly detection.
\newblock In \emph{Proceedings of the IEEE/CVF Conference on Computer Vision and Pattern Recognition}, 22883--22892.

\bibitem[{Wang et~al.(2024{\natexlab{b}})Wang, Bai, Tan, Wang, Fan, Bai, Chen, Liu, Wang, Ge et~al.}]{qwen2-vl}
Wang, P.; Bai, S.; Tan, S.; Wang, S.; Fan, Z.; Bai, J.; Chen, K.; Liu, X.; Wang, J.; Ge, W.; et~al. 2024{\natexlab{b}}.
\newblock Qwen2-vl: Enhancing vision-language model's perception of the world at any resolution.
\newblock \emph{arXiv preprint arXiv:2409.12191}.

\bibitem[{Wei et~al.(2022)Wei, Wang, Schuurmans, Bosma, Xia, Chi, Le, Zhou et~al.}]{cot}
Wei, J.; Wang, X.; Schuurmans, D.; Bosma, M.; Xia, F.; Chi, E.; Le, Q.~V.; Zhou, D.; et~al. 2022.
\newblock Chain-of-thought prompting elicits reasoning in large language models.
\newblock \emph{Advances in neural information processing systems}, 35: 24824--24837.

\bibitem[{Wieler and Hahn(2007)}]{dagm}
Wieler, M.; and Hahn, T. 2007.
\newblock Weakly supervised learning for industrial optical inspection.
\newblock In \emph{DAGM symposium in}, volume~6, 11.

\bibitem[{Wu et~al.(2023)Wu, Gan, Chen, Wan, and Yu}]{mllm_survey}
Wu, J.; Gan, W.; Chen, Z.; Wan, S.; and Yu, P.~S. 2023.
\newblock Multimodal large language models: A survey.
\newblock In \emph{2023 IEEE International Conference on Big Data (BigData)}, 2247--2256. IEEE.

\bibitem[{Xu et~al.(2025)Xu, Lo, Safaei, Patel, and Dwivedi}]{anomaly-ov}
Xu, J.; Lo, S.-Y.; Safaei, B.; Patel, V.~M.; and Dwivedi, I. 2025.
\newblock Towards zero-shot anomaly detection and reasoning with multimodal large language models.
\newblock In \emph{Proceedings of the Computer Vision and Pattern Recognition Conference}, 20370--20382.

\bibitem[{Yang et~al.(2025)Yang, Li, Yang, Zhang, Hui, Zheng, Yu, Gao, Huang, Lv et~al.}]{qwen3}
Yang, A.; Li, A.; Yang, B.; Zhang, B.; Hui, B.; Zheng, B.; Yu, B.; Gao, C.; Huang, C.; Lv, C.; et~al. 2025.
\newblock Qwen3 technical report.
\newblock \emph{arXiv preprint arXiv:2505.09388}.

\bibitem[{Yin et~al.(2024)Yin, Fu, Zhao, Li, Sun, Xu, and Chen}]{mllm}
Yin, S.; Fu, C.; Zhao, S.; Li, K.; Sun, X.; Xu, T.; and Chen, E. 2024.
\newblock A survey on multimodal large language models.
\newblock \emph{National Science Review}, 11(12): nwae403.

\bibitem[{You et~al.(2022)You, Cui, Shen, Yang, Lu, Zheng, and Le}]{uniad}
You, Z.; Cui, L.; Shen, Y.; Yang, K.; Lu, X.; Zheng, Y.; and Le, X. 2022.
\newblock A unified model for multi-class anomaly detection.
\newblock \emph{Advances in Neural Information Processing Systems}, 35: 4571--4584.

\bibitem[{Zeng et~al.(2025)Zeng, Pang, Wang, and Yang}]{lr-iad}
Zeng, P.; Pang, F.; Wang, Z.; and Yang, A. 2025.
\newblock LR-IAD: Mask-Free Industrial Anomaly Detection with Logical Reasoning.
\newblock \emph{arXiv preprint arXiv:2504.19524}.

\bibitem[{Zhang et~al.(2024)Zhang, Yu, Dong, Li, Su, Chu, and Yu}]{mmllm}
Zhang, D.; Yu, Y.; Dong, J.; Li, C.; Su, D.; Chu, C.; and Yu, D. 2024.
\newblock Mm-llms: Recent advances in multimodal large language models.
\newblock \emph{arXiv preprint arXiv:2401.13601}.

\bibitem[{Zhou et~al.(2023)Zhou, Pang, Tian, He, and Chen}]{anomalyclip}
Zhou, Q.; Pang, G.; Tian, Y.; He, S.; and Chen, J. 2023.
\newblock Anomalyclip: Object-agnostic prompt learning for zero-shot anomaly detection.
\newblock \emph{arXiv preprint arXiv:2310.18961}.

\bibitem[{Zhu and Pang(2024)}]{inctrl}
Zhu, J.; and Pang, G. 2024.
\newblock Toward generalist anomaly detection via in-context residual learning with few-shot sample prompts.
\newblock In \emph{Proceedings of the IEEE/CVF conference on computer vision and pattern recognition}, 17826--17836.

\bibitem[{Zou et~al.(2022)Zou, Jeong, Pemula, Zhang, and Dabeer}]{visa}
Zou, Y.; Jeong, J.; Pemula, L.; Zhang, D.; and Dabeer, O. 2022.
\newblock Spot-the-difference self-supervised pre-training for anomaly detection and segmentation.
\newblock In \emph{European conference on computer vision}, 392--408. Springer.

\end{thebibliography}

\section{Supplementary Material}
We present additional data construction process, methodological details, implementation specifics, experiments, and visualizations in this Supplementary Material. The Supplementary Material is organized as follows:
\begin{itemize}
    \item Section A provides the construction process of our Expert-AD, including a detailed pipeline for CoT generation and subsequent data curation steps.
    \item Section B offers detailed supplementary information on the two-stage training process of our IAD-R1.
    \item Section C presents specific experimental configurations, ablation study analyses, and visualization results.
    \item Section D provides a comprehensive summary of IAD-R1, analyzing its limitations and future optimization directions.
\end{itemize}
\subsection{A. Expert-AD Dataset}
\subsubsection{A.1. Dataset Construction}
% COT监督修改机制，如何确保生成质量？
The field of industrial image anomaly detection faces the challenge of lacking high-quality prompt-tuning datasets, which severely limits the application effectiveness of Vision-Language Models (VLMs) in this downstream task. Although the Anomaly-Instruct-125k~\cite{anomaly-ov} dataset proposed by Anomaly-ov~\cite{anomaly-ov} contains substantial instruction-tuning data, it suffers from two critical limitations: first, these data are primarily sourced from web crawling, resulting in significant discrepancies with images from real industrial scenarios; second, these prompts only contain simple image descriptions and answers, lacking complete anomaly detection reasoning logic and failing to effectively guide models in establishing systematic detection thinking frameworks.

To address these issues, we construct Expert-AD (totaling 5.9K QA pairs)—a high-quality real industrial scenario tuning dataset incorporating CoT reasoning. To ensure that images in our dataset authentically reflect real industrial application scenarios, we select Real-IAD~\cite{real-iad} as the image data source for Expert-AD. Real-IAD provides a large-scale real industrial anomaly detection dataset covering 30 categories with various anomaly types, establishing a solid data foundation for our research.
\begin{figure}[htbp]
    \centering
    \includegraphics[width=1\linewidth]{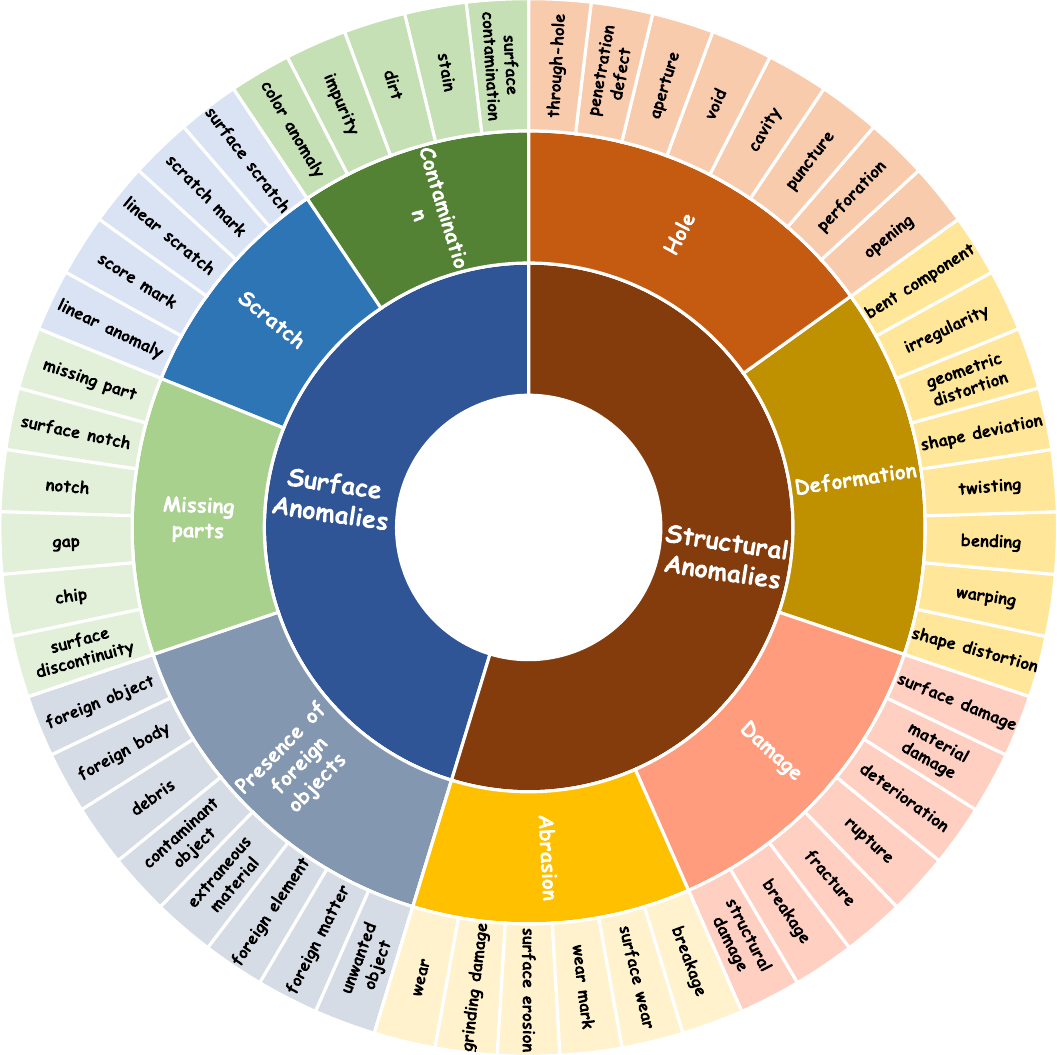}
    \caption{Detailed anomaly types in the Expert-AD dataset.}
    \label{anomaly_type}
\end{figure}

\begin{figure*}[htbp]
    \centering
    \includegraphics[width=1\linewidth]{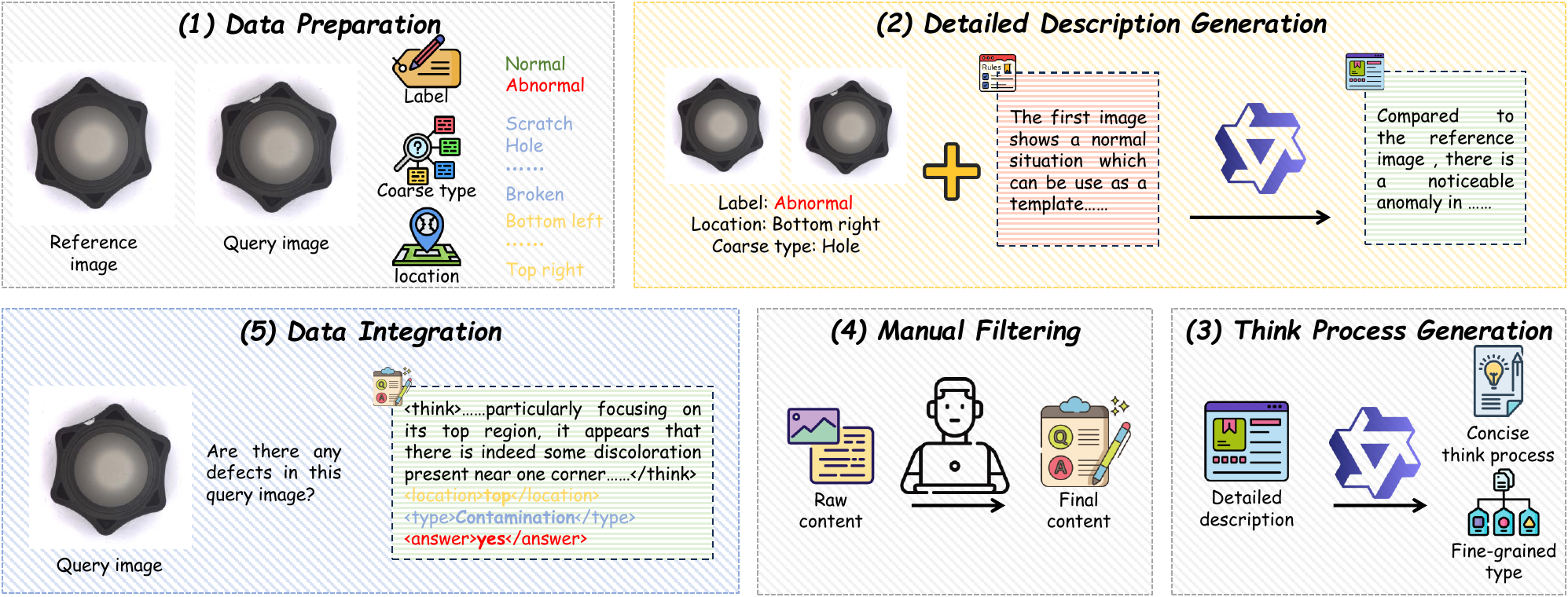}
    \caption{Data generation pipeline for Expert-AD dataset construction. The pipeline illustrates the complete workflow from raw data collection to high-quality training data, including preprocessing, CoT generation, quality control, and data filtering steps.}
    \label{datageneration}
\end{figure*}

Specifically, as shown in Figure \ref{datageneration} the generation process of the Expert-AD dataset can be summarized into the following steps:
\begin{itemize}
    \item Step 1: Data Preparation
In the data preparation phase, we first collect reference and query image pairs sourced from the Real-IAD dataset to ensure representativeness of real industrial scenarios. Simultaneously, we extract and organize relevant annotation information, including normal/abnormal labels, specific anomaly types (such as Scratch, Hole, Broken, etc.), coarse-grained type classifications, and precise anomaly location information (such as Bottom left, Top right, etc.). This step establishes the foundational data support for subsequent description generation and reasoning construction.
    \item Step 2: Detailed Description Generation
In the detailed description generation phase, we utilize Qwen-VL-Max~\cite{qwen-vl-max} to conduct in-depth analysis and description of images. We feed the query image along with its corresponding normal reference image to Qwen-VL-Max. For images containing anomalies, we highlight the anomalous regions and provide additional inputs including the original coarse-grained labels and anomaly location information derived from the mask annotations. The model then generates structured descriptions by comparing the query and reference images, producing detailed descriptive text such as ``Compared to the reference image, there is a noticeable anomaly in...".
    \item Step 3: Think Process Generation
The think process generation phase transforms detailed descriptions into structured reasoning chains while simultaneously generating fine-grained anomaly types. Through further model processing, lengthy detailed descriptions are distilled into concise and logically clear thinking processes. Leveraging chain-of-thought (CoT)~\cite{cot} responses and predefined multi-level anomaly types, this step generates more precise and diverse fine-grained anomaly types, enhancing data diversity to better align with practical industrial inspection requirements.
    \item Step 4: Manual Filtering
In the manual filtering phase, we conduct rigorous quality control on the generated content. Human experts review the raw generated content to filter out low-quality, inconsistent, or inaccurate data entries, ensuring that only high-quality data proceeds to the final integration stage.
    \item Step 5: Data Integration
The final manual integration phase systematically integrates all generated components. We organize questions, thinking processes, anomaly locations, anomaly types, and final answers according to standardized formats, constructing complete CoT structures based on whether images are normal/abnormal:
\begin{multline*}
    N = \langle think \rangle \text{...} \langle /think \rangle \langle answer \rangle \text{...} \langle /answer \rangle
\end{multline*}
\begin{multline*}
    A = \langle think \rangle \text{...} \langle /think \rangle \langle location \rangle \text{...} \langle /location \rangle\\
    \langle type \rangle \text{...} \langle /type \rangle \langle answer \rangle \text{...} \langle /answer \rangle
\end{multline*}
\end{itemize}
Finally, we finalize the data formatting and structure validation to ensure consistency across all entries, ultimately forming the high-quality Expert-AD training dataset.

\subsubsection{A.2. Dataset Analysis}
We randomly sample from the Perception Activation Supervised Fine-Tuning (PA-SFT) stage training samples of the Expert-AD dataset at different ratios, using 20\%, 40\%, 60\%, 80\%, and 100\% of the data for PA-SFT stage training to validate the dataset's effectiveness and explore optimal data scale configurations.

As shown in Table 6, different types of anomaly detection tasks exhibit significant differences in data volume requirements. For complex industrial component datasets (MVTec~\cite{mvtec}, VisA~\cite{visa}, MPDD~\cite{mpdd}), model performance shows a clear upward trend with increasing data volume. Taking the 0-shot setting as an example, the accuracy of the MVTec dataset improves from 85.0\% with 20\% data to 86.7\% with full data, VisA improves from 74.8\% to 78.0\%, and MPDD improves from 66.5\% to 70.9\%. This indicates that anomaly detection in complex industrial components requires more training data to learn fine-grained defect patterns.

In contrast, surface texture datasets (DAGM~\cite{dagm}, DTD~\cite{dtd}, SDD~\cite{sdd}) achieve near-optimal performance with relatively less data. For example, DAGM and DTD achieve 93.2\% and 93.4\% accuracy respectively with only 20\% of the data, approaching the performance achieved with the full dataset. This suggests that surface texture anomalies have relatively obvious visual features, enabling models to activate detection capabilities for such anomalies with fewer training samples.
This finding provides important guidance for practical applications: for surface texture anomaly detection tasks, relatively fewer high-quality CoT data can be used for training, while for complex industrial components, more comprehensive training data is recommended to ensure optimal performance.

\begin{figure}[htbp]
    \centering
    \includegraphics[width=1\linewidth]{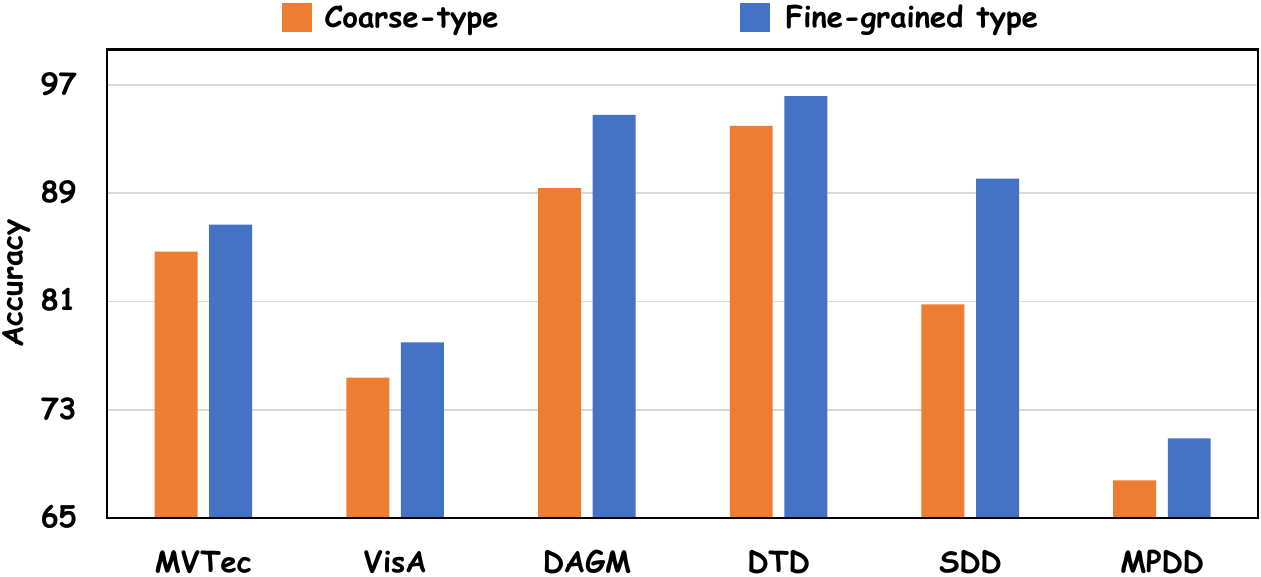}
    \caption{Performances of Coarse-type and Fine-grained-type on LLaVA-OneVision-SI-7B.}
    \label{ablation_type}
\end{figure}

\begin{table*}[htbp]
\centering
\begin{tabular}{ccccccccccc}
\toprule
\multirow{2}{*}{\textbf{Data Percentage}} & \multicolumn{7}{c}{\textbf{0-shot}} & \multicolumn{3}{c}{\textbf{1-shot}} \\
\cmidrule(lr){2-8} \cmidrule(lr){9-11}
& \textbf{MVTec} & \textbf{DAGM} & \textbf{DTD} & \textbf{SDD} & \textbf{MPDD} & \textbf{VisA} & \textbf{Average} & \textbf{MVTec} & \textbf{VisA} & \textbf{Average} \\
\midrule
20\% & 85.0 & 93.2 & 93.4 & \textbf{91.7} & 66.5 & 74.8 & 84.1 & 77.1 & 66.4 & 71.8 \\
40\% & 85.5 & 93.2 & 94.7 & 88.8 & 66.7 & 74.8 & 84.0 & \textbf{80.6} & 69.9 & 75.3 \\
60\% & \textbf{88.1} & 91.4 & 94.6 & 85.4 & 68.7 & 74.9 & 83.9 & 80.5 & 67.3 & 73.9 \\
80\% & 86.5 & 93.2 & 92.7 & 80.0 & 70.5 & 77.9 & 83.5 & 78.9 & 69.4 & 74.2 \\
100\% & 86.7 & \textbf{94.8} & \textbf{96.2} & 90.1 & \textbf{70.9} & \textbf{78.0} & \textbf{86.1} & 80.0 & \textbf{71.5} & \textbf{75.8} \\
\bottomrule
\end{tabular}%
\caption{Scale ablation study on Expert-AD dataset. Performance comparison using different percentages of training data (20\%, 40\%, 60\%, 80\%, 100\%) across six industrial anomaly detection datasets in 0-shot and 1-shot settings.}
\label{data_percentage}
\end{table*}

\subsubsection{A.3. Exploration of Fine-grained Anomaly Type}
% 这个地方放一个对比图/表，说明我们构建的细粒度的异常类型是有效的（可以涨点）
During the construction of the Expert-AD dataset, we further refined the original 8 coarse-grained anomaly types into more precise fine-grained types. To validate the effectiveness of this type refinement strategy, we designed comparative experiments to evaluate the impact of different granularity anomaly type annotations on model performance. We conducted PA-SFT stage training using coarse-type and fine-grained type annotations respectively, keeping all other training configurations identical. As shown in Figure \ref{ablation_type}, training data with fine-grained type annotations achieves better performance across all datasets. This improvement primarily stems from the fact that coarse-grained anomaly classifications are overly broad, grouping anomalies with different visual characteristics into the same category, failing to provide sufficiently precise supervisory signals for the model. In contrast, our fine-grained anomaly type system provides more accurate anomaly descriptions, enabling models to learn unique features of different anomaly types and establish more precise anomaly-language mapping relationships. This result validates the effectiveness of anomaly type refinement during Expert-AD dataset construction.

% 增加数据量分析，数据比例分析，数据集对比
\subsection{B. IAD-R1 Two-Stage Post-Training Process}
\subsubsection{B.1. IAD-R1 Algorithm}
% 这个地方对算法流程图进行描述，把算法流程图中的变量陈述一下
The complete two-stage training process of IAD-R1 is summarized in Algorithm \ref{alg_iad-r1}, which integrates both PA-SFT and Structured Control Group Relative Policy Optimization (SC-GRPO) stages into a unified training framework.
\begin{algorithm}[!htpb]
\caption{IAD-R1 Training Process}
\label{alg_iad-r1}
\textbf{Input}: Dataset $\mathcal{D}_{\text{Expert-AD}} = \{(I_i, p ,O_i)\}_{i=1}^N$, VLM $\pi_{\theta_0}$ \\
\textbf{Parameter}: KL penalty $\beta$, learning rate $\alpha$, clip $\varepsilon$, epochs $E$\\
\textbf{Output}: Optimized model $\pi_{\theta^*}$
\begin{algorithmic}[1]
\STATE \textbf{// Stage 1: PA-SFT}
\STATE Initialize $\pi_\theta \leftarrow \pi_{\theta_0}$
\FOR{each $(I, p, O) \in \mathcal{D}_{\text{Expert-AD}}$}
    \IF{$y = \text{normal}$}
        \STATE $O \leftarrow \langle\text{think}\rangle T \langle/\text{think}\rangle \langle\text{answer}\rangle A \langle/\text{answer}\rangle$
    \ELSE
        \STATE $O \leftarrow \langle\text{think}\rangle T \langle/\text{think}\rangle \langle\text{location}\rangle L \langle/\text{location}\rangle$
        \STATE \quad $\langle\text{type}\rangle t \langle/\text{type}\rangle \langle\text{answer}\rangle A \langle/\text{answer}\rangle$
    \ENDIF
\ENDFOR
\STATE Optimize: $\mathcal{L} = -\mathbb{E} \left[\sum_{i=1}^L \log \pi_\theta(o_i|I, p, o_{<i})\right]$
\STATE $\pi_{\text{PA-SFT}} \leftarrow \pi_\theta$
\STATE 
\STATE \textbf{// Stage 2: SC-GRPO}
\STATE Initialize $\pi_\theta \leftarrow \pi_{\text{PA-SFT}}$, $\pi_{\text{ref}} \leftarrow \pi_{\text{PA-SFT}}$
\FOR{epoch $= 1$ to $E$}
    \STATE Sample batch $\mathcal{B} = \{(I_i, p_i)\}_{i=1}^M$ from $\mathcal{D}_{\text{Expert-AD}}$
    \FOR{each $(I, p) \in \mathcal{B}$}
        \STATE Generate $G$ responses: $\{o_i^{(j)}\}_{j=1}^G \sim \pi_\theta(\cdot|I, p)$
        \FOR{$j = 1$ to $G$}
            \STATE $R_{\text{con}} \leftarrow \textsc{ConsistencyReward}(o_i^{(j)}, y_i)$
            \STATE $R_{\text{acc}} \leftarrow \textsc{AccuracyReward}(o_i^{(j)}, y_i)$
            \IF{$y_i = \text{anomalous}$}
                \STATE $R_{\text{loc}} \leftarrow \textsc{LocationReward}(o_i^{(j)}, l_i)$
                \STATE $R_{\text{type}} \leftarrow \textsc{TypeReward}(o_i^{(j)}, t_i)$
            \ENDIF
            \STATE $R_{\text{total}} \leftarrow R_{\text{con}} + R_{\text{acc}} + \mathbb{I}\{y_i=\text{anomalous}\} \cdot (R_{\text{loc}} + R_{\text{type}})$
            \STATE $A_j \leftarrow \frac{R_{\text{total}} - \mu_R}{\sigma_R}$ \COMMENT{Normalize}
        \ENDFOR
    \ENDFOR
    \STATE Compute weights: $\rho_\theta(o_i^{(j)}) = \frac{\pi_\theta(o_i^{(j)}|I, p)}{\pi_{\text{ref}}(o_i^{(j)}|I, p)}$
    \STATE Compute KL: $D_{\text{KL}} = \mathbb{E}[\log \pi_\theta(o|I,p) - \log \pi_{\text{ref}}(o|I,p)]$
    \STATE Update: $\theta \leftarrow \theta + \alpha\nabla_\theta \mathcal{J}$
    \STATE where $\mathcal{J} = \mathbb{E}[\min(\rho_\theta A_j, \text{clip}(\rho_\theta, 1\pm\varepsilon)A_j)] - \beta D_{\text{KL}}$
\ENDFOR
\STATE \textbf{return} $\pi_{\theta^*}$
\end{algorithmic}
\end{algorithm}

% 这个地方增加算法流程图
\subsubsection{B.2. Perception Activation Supervised Fine-Tuning}
In the PA-SFT stage, we activate the model's anomaly perception potential through full-parameter supervised fine-tuning on the Expert-AD dataset.
The core objective of this stage is to enable the model to master structured anomaly detection reasoning patterns and ensure logical consistency between the reasoning process and final judgment results. 
Training samples adopt a triplet format $(I, p, O)$, where $I$ represents the industrial image, $p$ denotes the text prompt, and $O$ is the conditional output sequence. Through high-quality CoT data training, the model learns to identify anomaly patterns from visual features and generates outputs with different structures based on image content: 

(1) Normal images: $O=(T,A)$, containing the CoT reasoning process $T$ and final answer $A$; (2) Anomalous images: $O = (T, L, t, A)$, containing the CoT reasoning process $T$, anomaly location $L$, anomaly type $t$, and final answer $A$. This training approach ensures that the model can produce logically coherent structured outputs under different anomaly states, avoiding contradictions between the reasoning process and final conclusions. The training objective function is defined as maximizing the probability of generating the conditional output sequence given the image and prompt conditions:
{\small
\begin{equation}
    \mathcal{L}_{\text{PA-SFT}} = -\mathbb{E}_{(I,p,O) \sim \mathcal{D}_{\text{Expert-AD}}} \sum_{i=1}^{L} \log \pi_\theta(o_i | I, p, o_{<i}),
\end{equation}
}where $D_{Expert-AD}$ is the Expert-AD dataset, $o_i$ is the $i$-th token in the output sequence $O$, $L$ is the sequence length, and $\pi_\theta$ is the model parameter distribution. The trained model serves as the initialization model $\pi_{PA-SFT}$ for the SC-GRPO stage.

\subsubsection{B.3. Structured Control Group Relative Policy Optimization}
We utilize the model trained through PA-SFT as the initial policy model and apply reinforcement learning strategies to further optimize model performance. Traditional Proximal Policy Optimization (PPO)~\cite{ppo} algorithms require training value function networks of comparable scale to the policy model, which not only introduces significant memory and computational burdens but also increases training complexity. In contrast, Group Relative Policy Optimization (GRPO)~\cite{grpo} performs intra-group relative advantage estimation on rewards from multiple sampled outputs, avoiding the need for additional value function network training and significantly reducing computational resource consumption and implementation complexity. Targeting the structured output characteristics and multi-dimensional evaluation requirements of industrial anomaly detection tasks, we perform task-specific optimization of the GRPO algorithm by designing specialized multi-dimensional reward function mechanisms. This mechanism not only ensures output format consistency but also considers detection accuracy, enabling the reinforcement learning process to better adapt to the conditional output requirements of anomaly detection.
\\
\textbf{Action Group Sampling Strategy.}
For each input state $s = (I, p)$, where $I$ represents the industrial image and $p$ denotes the text prompt, SC-GRPO samples a group of actions $\{o_1, o_2, \ldots, o_G\}$ from the current policy $\pi_\theta$ (initialized by $\pi_{PA-SFT}$):
{\small
\begin{equation}
    o_i \sim \pi_\theta(o | I, p), \quad i = 1, 2, \ldots, G,
\end{equation}
}where $G$ is the group size, and each $o_i$ represents a complete structured output sequence. This diversified sampling strategy helps explore different reasoning pathways and prevents the model from prematurely converging to suboptimal solutions.
\\
\textbf{Reward Function Design.}
In order to improve the anomaly understanding ability of IAD-R1, SC-GRPO employs a comprehensive reward function that integrates four specialized components: consistency reward ($R_{con}$), answer accuracy reward ($R_{acc}$), location accuracy reward ($R_{loc}$), and type accuracy reward ($R_{type}$). The total reward for anomalous samples is computed as:
{\small
\begin{equation}
    R_{\text{SC-GRPO}}(o_i) = R_{con} + R_{acc} + \mathbb{I}\{y_i = \text{anomalous}\} \cdot (R_{loc} + R_{type}),
\end{equation}
}where $\mathbb{I}\{y_i = \text{anomalous}\}$ is an indicator function that equals 1 for anomalous samples and 0 for normal samples. For normal samples, only consistency and accuracy rewards are applied, while anomalous samples additionally receive location and  
type rewards when predictions are correct. Detailed descriptions of each reward component and their mathematical formulations are provided in the main paper.
\\
\textbf{Relative Advantage Computation.}
We adopt an intra-group reward normalization strategy to compute relative advantages, avoiding the training overhead of value function networks in PPO~\cite{ppo}:
\begin{equation}
    A_i = \frac{R_{\text{SC-GRPO}}(o_i) - \text{mean}\{R_{\text{SC-GRPO}}(o_j)\}_{j=1}^G}{\text{std}\{R_{\text{SC-GRPO}}(o_j)\}_{j=1}^G}.
\end{equation}
\\
\textbf{Policy Update Mechanism.}
SC-GRPO adopts a PPO-style clipped objective function combined with KL divergence constraints for policy updates:
{\small
\begin{multline}
    \mathcal{J}_{\text{SC-GRPO}}(\theta) = \\
    \mathbb{E}_{(I,p),\{o_i\} \sim \pi_{\theta_{\text{old}}}} \left[ \frac{1}{G} \sum_{i=1}^G \mathcal{L}_{\text{CLIP}}(o_i, (I,p), \theta) \right] \\
    - \beta \mathcal{D}_{\text{KL}}(\pi_\theta \| \pi_{\text{ref}}),
\end{multline}
}where $\mathcal{L}_{CLIP}$ is the PPO clipped loss, $\beta$ is the KL regularization coefficient, and $\pi_{ref}$ is the reference policy.

\subsection{C. Experimental Details and Analysis}
\subsubsection{C.1. Experiment Setup}
In this section, we provide detailed descriptions of the training and testing configurations, backbone architectures, baseline method, and specific implementation details.
\\
\textbf{Training Setup.} We adopt a zero-shot training paradigm, where the training data is entirely sourced from our constructed auxiliary dataset Expert-AD, containing no samples from the target testing domains.
\\
\textbf{Evaluation Datasets.} To evaluate IAD-R1's detection capabilities in industrial anomaly detection, we select the complete test sets from six representative industrial anomaly detection datasets as our target testing datasets, encompassing two major categories: industrial components (MVTec-AD~\cite{mvtec}, VisA~\cite{visa}, MPDD~\cite{mpdd}) and surface textures (DAGM~\cite{dagm}, DTD~\cite{dtd}, SDD~\cite{sdd}), aimed at simulating complex real-world industrial production scenarios.
\\
\textbf{Testing Setup.} We conduct evaluations on the target domains under two settings:
(1) Zero-shot testing (0-shot): The model receives only text prompts and test images as input, outputting model reasoning and anomaly detection results.
(2) One-shot testing (1-shot): In addition to text prompts and test images, the model additionally receives a normal reference image from the same product category as contextual information, with outputs consisting of model reasoning and anomaly detection results. For the MVTec and VisA test sets that support 1-shot settings, we adopt the same normal reference image configuration as MMAD~\cite{mmad}, providing corresponding normal reference samples for each test image.
\\
\textbf{Evaluate Metric.}
In industrial anomaly detection scenarios, correctly identifying both normal and abnormal samples is of equal importance. Due to the class imbalance commonly present in anomaly detection datasets, overall accuracy is prone to evaluation bias—even when a model consistently outputs predictions for a single category, it may still achieve high accuracy, which fails to truly reflect the model's actual performance on both normal and abnormal sample categories.
Therefore, we adopt average accuracy as the primary evaluation metric, which provides a more balanced and reliable performance measure by computing the arithmetic mean of normal sample accuracy and abnormal sample accuracy. 

Specifically, for models trained through the IAD-R1 framework, we extract the final prediction results from the $\langle answer \rangle \text{...} \langle /answer \rangle$ tags in their structured output format for comparison with ground truth labels; for other baseline models, we directly obtain their raw output answers for evaluation comparison. Detailed prompts can be found in Table \ref{prompt_design}.
\\
\textbf{Backbone Selection.}
To validate the generalization performance of IAD-R1, we select pre-trained multimodal large models with different architectures and parameter scales as backbones for experimentation, including Qwen2-VL-2B~\cite{qwen2-vl}, Qwen2.5-VL-3B~\cite{qwen2.5-vl}, Qwen2.5-VL-7B~\cite{qwen2.5-vl}, LLaVA-1.5-7B~\cite{llava-1.5}, LLaVA-1.6-8B~\cite{llava-next}, LLaVA-OneVision-SI-0.5B~\cite{llava-ov-si}, and LLaVA-OneVision-SI-7B~\cite{llava-ov-si}. These models span a parameter range from 0.5B to 8B, encompassing two mainstream multimodal large language model series—Qwen and LLaVA—enabling comprehensive validation of the IAD-R1 framework's applicability and effectiveness across different model architectures and scales.
\\
\textbf{Comparison Method.}
To comprehensively evaluate the performance of IAD-R1, we select multiple types of baseline models for comparative experiments, specifically including:

Commercial Large Models: We select state-of-the-art commercial multimodal large language models, including GPT-4o-mini-2024-07-18~\cite{gpt-4o}, GPT-4o-2024-08-06~\cite{gpt-4o}, GPT-4.1-nano-2025-04-14~\cite{gpt4.1}, GPT-4.1-mini-2025-04-14~\cite{gpt4.1}, GPT-4.1-2025-04-14~\cite{gpt4.1}, and Claude-Sonnet-4-2025-05-14~\cite{claude4}, which represent the highest level of commercial VLMs.

Open-Source Large Models: In addition to the models used as backbones, we introduce other mainstream open-source VLMs for comparison, including InternVL-2.5-4B~\cite{internvl2.5}, LLaVA-1.5-13B~\cite{llava-1.5}, LLaVA-1.6-34B~\cite{llava-next}, and Qwen2.5-VL-Instruct-72B~\cite{qwen2.5-vl}, covering different parameter scales and architectural designs.

Specialized Anomaly Detection Large Models: We compare against three methods specifically designed for anomaly detection tasks: (1) AnomalyGPT~\cite{anomalygpt}, an industrial anomaly detection approach based on large vision-language models that generates training data by simulating anomalous images and employs an image decoder with prompt learning for fine-tuning, enabling direct assessment of anomaly presence and location without manual threshold adjustments; (2) Anomaly-ov~\cite{anomaly-ov}, a specialist visual assistant for zero-shot anomaly detection and reasoning that employs a LookTwice Feature Matching mechanism to adaptively select abnormal visual tokens and constructs the Anomaly-Instruct-125k instruction tuning dataset for model training; (3) AnomalyR1~\cite{anomalyr1}, an industrial anomaly detection framework based on the multimodal large language model VLM-R1 that integrates Group Relative Policy Optimization (GRPO) with Reasoned Outcome Alignment Metric (ROAM) to achieve end-to-end anomaly detection.
\\
\textbf{Implementation Details.}
All training and testing were conducted on a server equipped with 4 $\times$ Nvidia RTX A100 80G GPUs. During the training process, we employed optimization techniques including vLLM, DeepSpeed, and Flash Attention 2 to enhance training efficiency and memory utilization. The entire training consists of two stages: the PA-SFT supervised fine-tuning stage trained for 2 epochs to ensure the model acquires fundamental anomaly perception capabilities while avoiding overfitting; the SC-GRPO reinforcement learning stage trained for 1 epoch to further enhance the model's anomaly detection and reasoning abilities through reward mechanisms. Both stages utilized bf16 mixed-precision training to accelerate the training process. During model evaluation, we employed the vLLM inference framework to accelerate the testing pipeline with an inference batch size of 4, significantly improving evaluation efficiency while maintaining inference quality.
%详细的提示词设计
\subsubsection{C.2. Prompt Design}
To ensure consistent and effective prompting strategies across different experimental phases and model architectures, we design targeted instruction templates. Table \ref{prompt_design} presents the comprehensive prompt configurations used across different experimental phases, including training, testing, ablation studies, and baseline method reproduction. In the testing phase, we use different prompts for different evaluation settings (0-shot and 1-shot) to guide the model's anomaly detection outputs. For baseline method reproduction, we carefully design prompts for AnomalyGPT, Anomaly-OV, and Anomaly-R1 that closely follow their original paper specifications and code implementations to ensure fair and accurate comparison.

\subsubsection{C.3. Detailed Ablation Results}
% 这个地方需要和正文保持一致，应该是数据制备方式和将函数设计策略。
To comprehensively validate the effectiveness of each component in the IAD-R1 framework, we conduct detailed ablation studies presenting complete results across all datasets.
\\
\textbf{PA-SFT Stage Ablation.} To verify the effectiveness of our Expert-AD dataset and process-aware supervised fine-tuning, we compare different data preparation strategies during the PA-SFT stage. As shown in Table \ref{appendix_ablation_sft}, we evaluate three data configurations: (1) ``Base" - training without any industrial anomaly detection data; (2) ``Original" - fine-tuning with the same images from Real-IAD but using direct answers without reasoning processes; (3) ``Expert-AD" - fine-tuning with our Expert-AD dataset containing complete CoT reasoning structures.
\\
\textbf{SC-GRPO Stage Ablation.} To verify the importance of our multi-dimensional reward design, we compare different optimization strategies on models that have undergone PA-SFT training. As shown in Table \ref{appendix_ablation_grpo}, we evaluate three strategies: (1) ``Fine-tuning" - the fine-tuned PA-SFT model without further optimization; (2) ``Original" - using only answer correctness as the reward signal; (3) ``SC-GRPO" - using our proposed multi-dimensional reward approach.

\subsubsection{C.4. Ablation Study on Fine-tuning Strategies}
\begin{figure}[!htbp]
    \centering
    \includegraphics[width=1\linewidth]{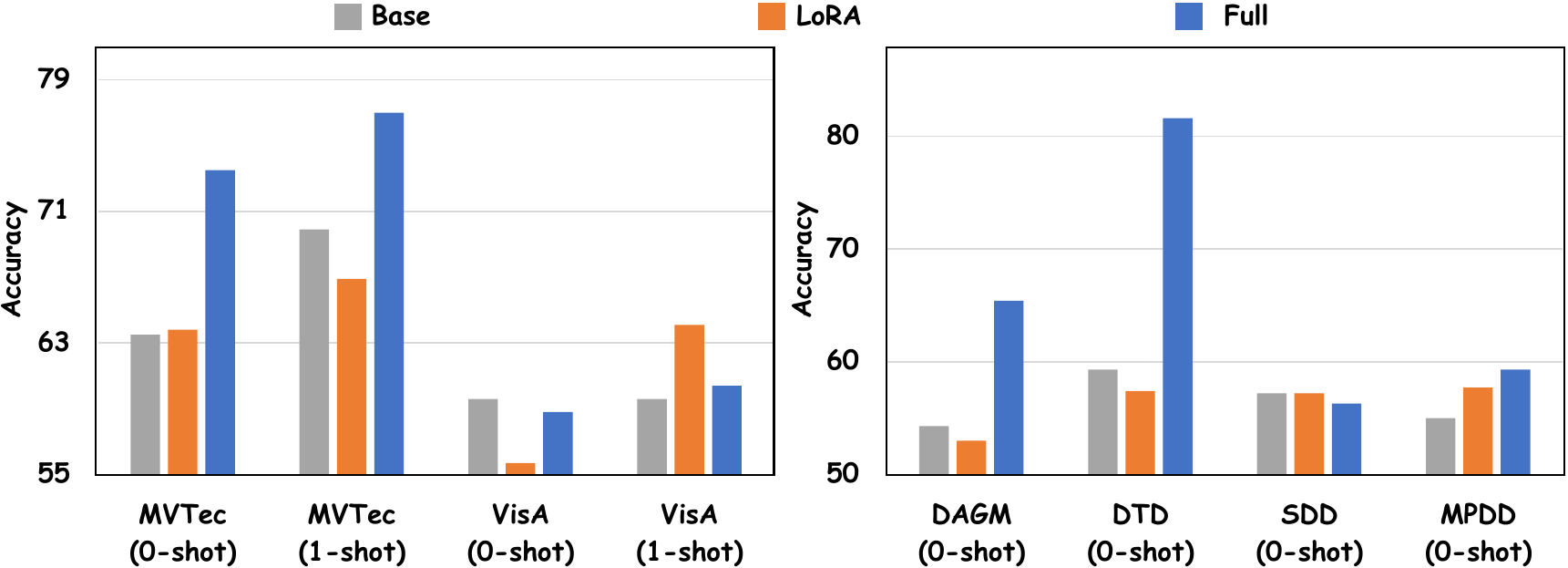}
    \caption{Performance comparison of different fine-tuning strategies on Qwen2-VL-Instruct-2B. Results show performance across different datasets in 0-shot and 1-shot settings for Base model, LoRA, and Full parameter fine-tuning.}
    \label{fine-tuning}
\end{figure}

% 这个地方的实验结果、实验数据需要确认具体数值；表格需要重新绘制（lora和Full）这个地方使用的模型是qwen2B
To validate the effectiveness of full parameter fine-tuning compared to parameter-efficient fine-tuning methods, we compare three different fine-tuning strategies: Base model (no fine-tuning), LoRA~\cite{lora} fine-tuning, and full parameter fine-tuning (Full). Experiments are conducted on the same Expert-AD dataset with consistent training configurations to ensure fair comparison.
As shown in Figure \ref{fine-tuning}, full parameter fine-tuning demonstrates significant advantages in the vast majority of testing scenarios. On the MVTec dataset, full parameter fine-tuning achieves 73.5\% accuracy in the 0-shot setting, improving by 9.7\% compared to LoRA fine-tuning's 63.8\%; in the 1-shot setting, it reaches 77.0\%, significantly outperforming LoRA's 66.9\%. On the DTD dataset, full parameter fine-tuning achieves 81.6\% accuracy in the 0-shot setting, markedly superior to LoRA's 57.4\%. Overall, full parameter fine-tuning shows varying degrees of performance improvement over LoRA fine-tuning across all datasets.
These results validate that industrial anomaly detection tasks require models to learn new vision-language mapping relationships and domain-specific knowledge. Full parameter fine-tuning can better reshape the model's knowledge structure, achieving effective transformation from general multimodal understanding to specialized anomaly detection. In contrast, LoRA's limited parameter update scope makes it difficult to adequately adapt to such cross-domain task transformation requirements. Therefore, we adopt the full parameter fine-tuning strategy in the IAD-R1 framework.

\begin{table*}[htbp]
\centering
\begin{tabular}{cccccccccccc}
\toprule
\multirow{2}{*}{\textbf{Method}} & \multicolumn{7}{c}{\textbf{0-shot}} & \multicolumn{3}{c}{\textbf{1-shot}} \\
\cmidrule(lr){2-8} \cmidrule(lr){9-11}
& \textbf{MVTec} & \textbf{DAGM} & \textbf{DTD} & \textbf{SDD} & \textbf{MPDD} & \textbf{VisA} & \textbf{Average} & \textbf{MVTec} & \textbf{VisA} & \textbf{Average} \\
\midrule
\multicolumn{11}{l}{\textbf{LLaVA-OneVision-SI-0.5B}} \\
\midrule
Anomaly-OV & 50.0 & 50.0 & 53.8 & 50.0 & 50.0 & 50.0 & 50.6 & 50.0 & 50.0 & 50.0 \\
IAD-R1 & \textbf{81.0} & \textbf{93.3} & \textbf{95.5} & \textbf{88.6} & \textbf{69.4} & \textbf{74.9} & \textbf{83.8} & \textbf{70.8} & \textbf{66.9} & \textbf{68.9} \\
\midrule
\multicolumn{11}{l}{\textbf{Qwen2.5-VL-Instruct-3B}} \\
\midrule
Anomaly-R1 & 69.4 & 56.7 & 61.0 & 57.6 & 56.0 & 59.8 & 60.1 & 69.4 & 62.2 & 65.8 \\
IAD-R1 & \textbf{77.6} & \textbf{85.2} & \textbf{89.1} & \textbf{83.4} & \textbf{59.2} & \textbf{69.8} & \textbf{77.4} & \textbf{78.6} & \textbf{74.1} & \textbf{76.4} \\
\midrule
\multicolumn{11}{l}{\textbf{AnomalyGPT (Vicuna-7B)}} \\
\midrule
Train on VisA & 46.6 & 49.6 & 64.1 & 49.5 & 54.2 & 57.3 & 53.6 & 74.6 & 74.0 & 74.3 \\
Train on MVTec & 46.6 & 60.4 & 56.9 & 42.2 & 45.8 & 57.3 & 51.5 & 74.6 & 74.0 & 74.3 \\
\midrule
\multicolumn{11}{l}{\textbf{LLaVA-OneVision-SI-7B}} \\
\midrule
Anomaly-OV & 74.3 & 77.5 & 90.7 & 88.7 & 70.3 & 71.8 & 78.9 & 70.3 & 63.9 & 67.1 \\
IAD-R1 & \textbf{86.7} & \textbf{94.8} & \textbf{96.2} & \textbf{90.1} & \textbf{70.9} & \textbf{78.0} & \textbf{86.1} & \textbf{80.0} & \textbf{71.5} & \textbf{75.8} \\
\bottomrule
\end{tabular}%
\caption{Detailed performance comparison of different methods across various settings.}
\label{anomaly_model_comparison}
\end{table*}

\begin{table*}[htbp]
\centering
\begin{tabular}{p{4cm}p{12cm}} %这个地方的命令可以使用吗？这个必须要确定好啊
\toprule
\textbf{Phase} & \textbf{Instruction/Prompt} \\
\midrule
GRPO Training & 
``You are an expert in detecting defects in image. Your task is to detect if there are any defects in the test image.\textbackslash n\texttt{\{Question\}}" \\
\midrule
SC-GRPO Training & 
``You are an expert in detecting defects in image. Your task is to detect if there are any defects in the test image.\textbackslash n\texttt{\{Question\}}" \\
\midrule
Using Base Model as Initial Policy Model & 
``You are an expert in detecting anomalies in image. Your task is to detect if there are any anomalies in the test image. If you find anomalies in the test image, structure your response with the following format: \texttt{<think>}[Your process of observation and reasoning is here]\texttt{</think>}\texttt{<location>}[The location of the anomaly in the image]\texttt{</location>}\texttt{<type>}[The type of anomaly in the image]\texttt{</type>}\texttt{<answer>}[Your final answer is here(yes or no)]\texttt{</answer>} If no anomalies are detected in the test image, structure your response with the following format: \texttt{<think>}[Your process of observation and reasoning is here]\texttt{</think>}\texttt{<answer>}[Your final answer is here(yes or no)]\texttt{</answer>}" \\
\midrule
IAD-R1 Test (0-shot) & 
``\texttt{\{image\}}Are there any defects in the test image?" \\
\midrule
IAD-R1 Test (1-shot) & 
``Following is 1 image of normal sample, which can be used as a template to compare the image being queried. \texttt{\{image\}} Following is image of test sample: \texttt{\{image\}} Are there any defects in the test image?" \\
\midrule
Base Model Test (0-shot) & 
``\texttt{\{image\}}Are there any defects in the query image? Please answer by yes or no." \\
\midrule
Base Model Test (1-shot) & 
``Following is 1 image of normal sample, which can be used as a template to compare the image being queried. \texttt{\{image\}} Following is image of test sample: \texttt{\{image\}} Are there any defects in the test image? Please answer by yes or no." \\
\midrule
AnomalyGPT Test (0-shot) & 
``Following is 1 image of normal sample,  which can be used as a template to compare the image being queried. \texttt{\{image\}} This is a photo of a object for anomaly detection, which should be round, without any damage, flaw, defect, scratch, hole or broken part. \texttt{\{iamge\}} Are there any defects in the test image? Please answer by yes or no." \\
\midrule 
AnomalyGPT Test (1-shot) &
``This is a photo of a object for anomaly detection, which should be round, without any damage, flaw, defect, scratch, hole or broken part. \texttt{\{iamge\}} Are there any defects in the test image? Please answer by yes or no." \\
\midrule
Anomaly-OV Test (0-shot) & ``Are there any defects for the object in the image? Please reply with 'Yes' or 'No'." \\
\midrule
Anomaly-OV Test (1-shot) & 
``Following is 1 image of normal sample, which can be used as a template to compare the image being queried. \texttt{\{image\}} Following is image of test sample: \texttt{\{image\}} Are there any defects for the object in the image? Please reply with 'Yes' or 'No'." \\
\midrule
Anomaly-R1 Test (0-shot) & 
``First output the thinking process in \texttt{<think>} \texttt{</think>} tags and then output the final answer letter in \texttt{<answer>} \texttt{</answer>} tags. \texttt{\{image\}} Are there any defects in the test image?" \\ 
\midrule
Anomaly-R1 Test (1-shot) & ``First output the thinking process in \texttt{<think>} \texttt{</think>} tags and then output the final answer letter in \texttt{<answer>} \texttt{</answer>} tags. Following is 1 image of normal sample,  which can be used as a template to compare the image being queried. \texttt{\{image\}} Following is image of test sample: \texttt{\{image\}} Are there any defects in the test image?" \\ 
\bottomrule
\end{tabular}
\caption{Prompt design for different phases and methods.}
\label{prompt_design}
\end{table*}

\subsubsection{C.5. Comparison of Anomaly Reasoning Model}
To comprehensively evaluate the effectiveness of IAD-R1, we select the latest VLM-based anomaly detection methods for comparison, including AnomalyGPT, Anomaly-OV, and AnomalyR1. To ensure experimental fairness and reproducibility, we strictly follow the experimental configurations described in the original papers, with specific reproduction details as follows:
\\
\textbf{Reproduction Configuration.}
\begin{itemize}
    \item AnomalyGPT: Since the original method uses both MVTec and VisA datasets as training and testing samples, which contradicts the zero-shot setting, we adopt a cross-validation strategy following zero-shot anomaly detection methods ~\cite{adaclip,anomalyclip}: when testing MVTec performance, we use the model trained on VisA, and when testing VisA performance, we use the model trained on MVTec, ensuring separation between training and testing data.
    \item AnomalyR1: We use the pre-trained model weights released by the authors, with prompt settings consistent with the official version. It should be noted that this method includes partial data from MVTec and VisA datasets during training.
    \item Anomaly-OV: We use the anomaly encoder weights released by the authors, with VLM selection of LLaVA-OneVision-SI-0.5B and LLaVA-OneVision-SI-7B, consistent with the original paper.
\end{itemize}
\textbf{Experimental Design.} To eliminate the influence of backbone selection on results, we apply our IAD-R1 strategy on the same backbones used by the aforementioned methods: LLaVA-OneVision-SI-0.5B, Qwen2.5-VL-Instruct-3B, and LLaVA-OneVision-SI-7B, ensuring fair comparison.
\\
\textbf{Results Analysis.} As shown in Table \ref{anomaly_model_comparison}, IAD-R1 achieves optimal performance across all configurations. In the 0-shot setting, IAD-R1 achieves average accuracies of 83.8\% (0.5B), 77.4\% (3B), and 86.1\% (7B). In the 1-shot setting, IAD-R1 also maintains significant advantages, achieving average accuracies of 68.9\% (0.5B), 76.4\% (3B), and 75.8\% (7B). These results validate the consistent advantages and strong generalization capabilities of the IAD-R1 framework across different model scales.

\begin{table*}[htbp]
\centering
\small
\begin{tabular}{ccccccccccc}
\toprule
\multirow{2}{*}{\textbf{Model}} & \multirow{2}{*}{\textbf{Parameter}} & \multirow{2}{*}{\textbf{Cold-start}} & \multicolumn{6}{c}{\textbf{0-shot}} & \multicolumn{2}{c}{\textbf{1-shot}} \\
\cmidrule(lr){4-9} \cmidrule(lr){10-11}
& & & \textbf{MVTec} & \textbf{VisA} & \textbf{DAGM} & \textbf{DTD} & \textbf{MPDD} & \textbf{SDD} & \textbf{MVTec} & \textbf{VisA} \\
\midrule
\multirow{2}{*}{LLaVA-OneVision-SI} & \multirow{2}{*}{0.5B} 
& Base & 50.0 & 49.9 & 50.0 & 51.6 & 50.0 & 50.0 & 50.0 & 50.0 \\
& & Fine-tuning & \textbf{81.0} & \textbf{74.9} & \textbf{93.3} & \textbf{95.5} & \textbf{69.4} & \textbf{88.6} & \textbf{70.8} & \textbf{66.8} \\
\multirow{2}{*}{Qwen2-VL-Instruct} & \multirow{2}{*}{2B} 
& Base & 67.6 & 57.8 & 55.6 & 61.3 & 62.8 & 60.3 & 69.8 & 61.0 \\
& & Fine-tuning & \textbf{77.3} & \textbf{67.7} & \textbf{73.8} & \textbf{84.0} & \textbf{69.5} & \textbf{62.8} & \textbf{78.5} & \textbf{69.7} \\
\bottomrule
\end{tabular}%
\caption{How about using base model as initial policy model?}
\label{initial_policy_model}
\end{table*}

\subsubsection{C.6. Visualization of Model Output}
\begin{figure*}[htbp]
    \centering
    \includegraphics[width=1\linewidth]{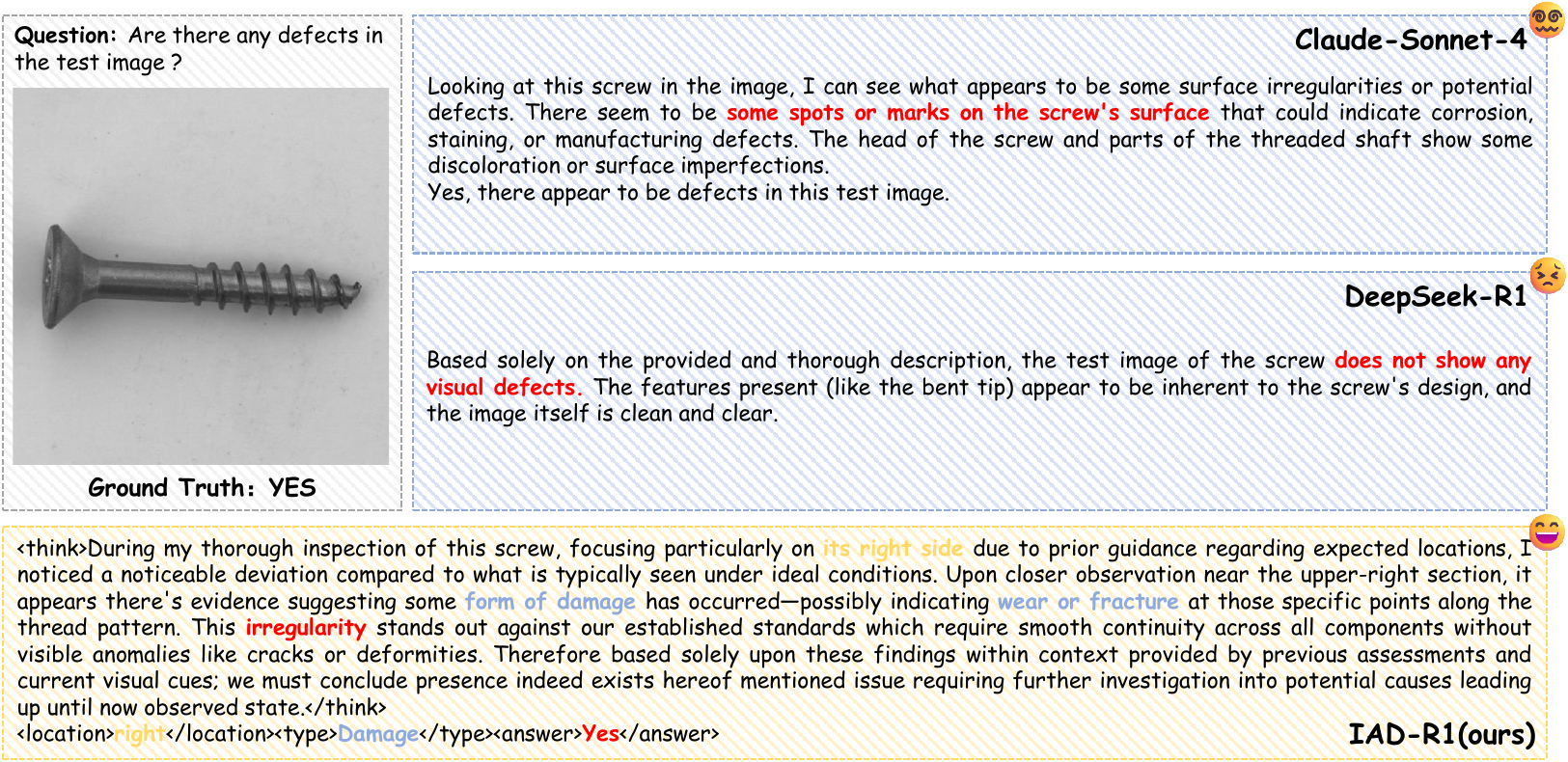}
    \caption{Different models' responses to a screw defect detection task.}
    \label{vis_screw}
\end{figure*}

\begin{figure*}[htbp]
    \centering
    \includegraphics[width=1\linewidth]{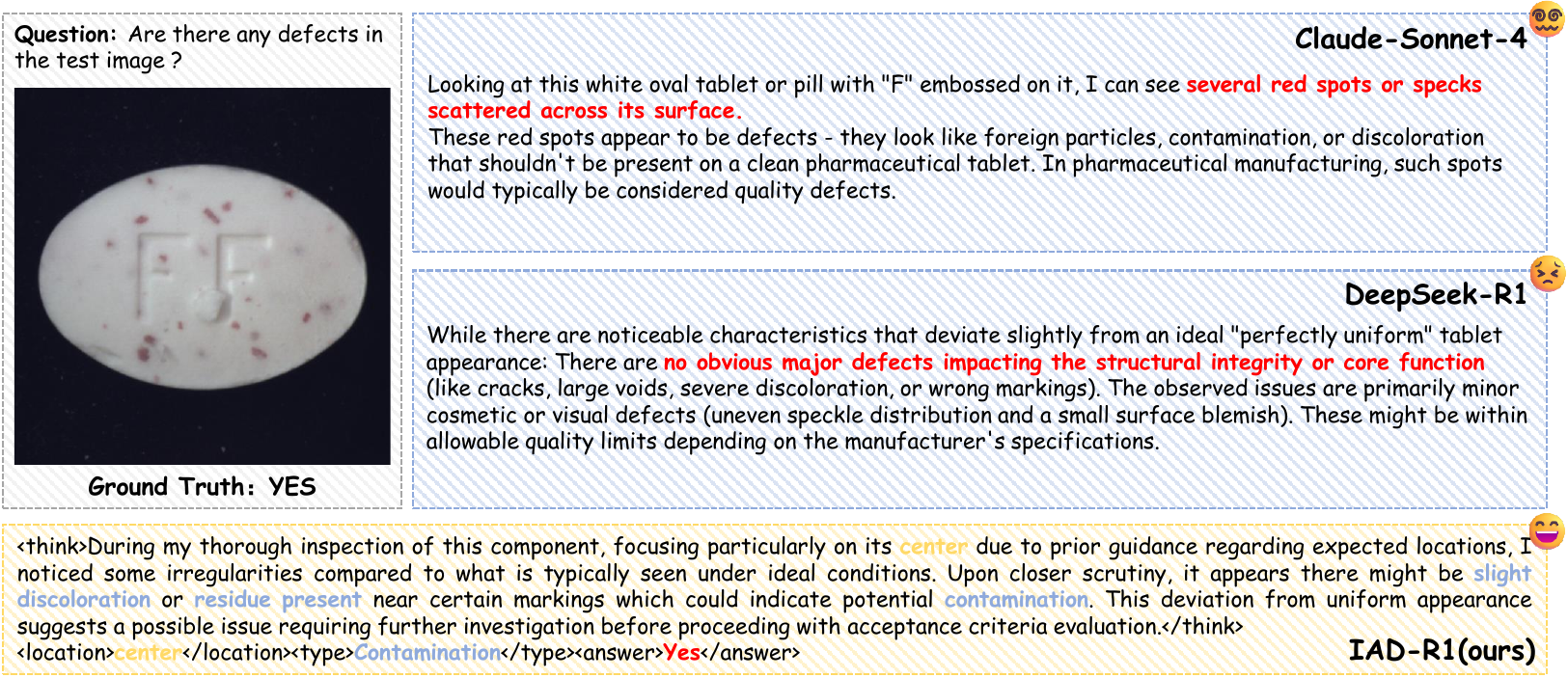}
    \caption{Different models' responses to a tablet defect detection task.}
    \label{vis_tablet}
\end{figure*}

\begin{figure*}[htbp]
    \centering
    \includegraphics[width=1\linewidth]{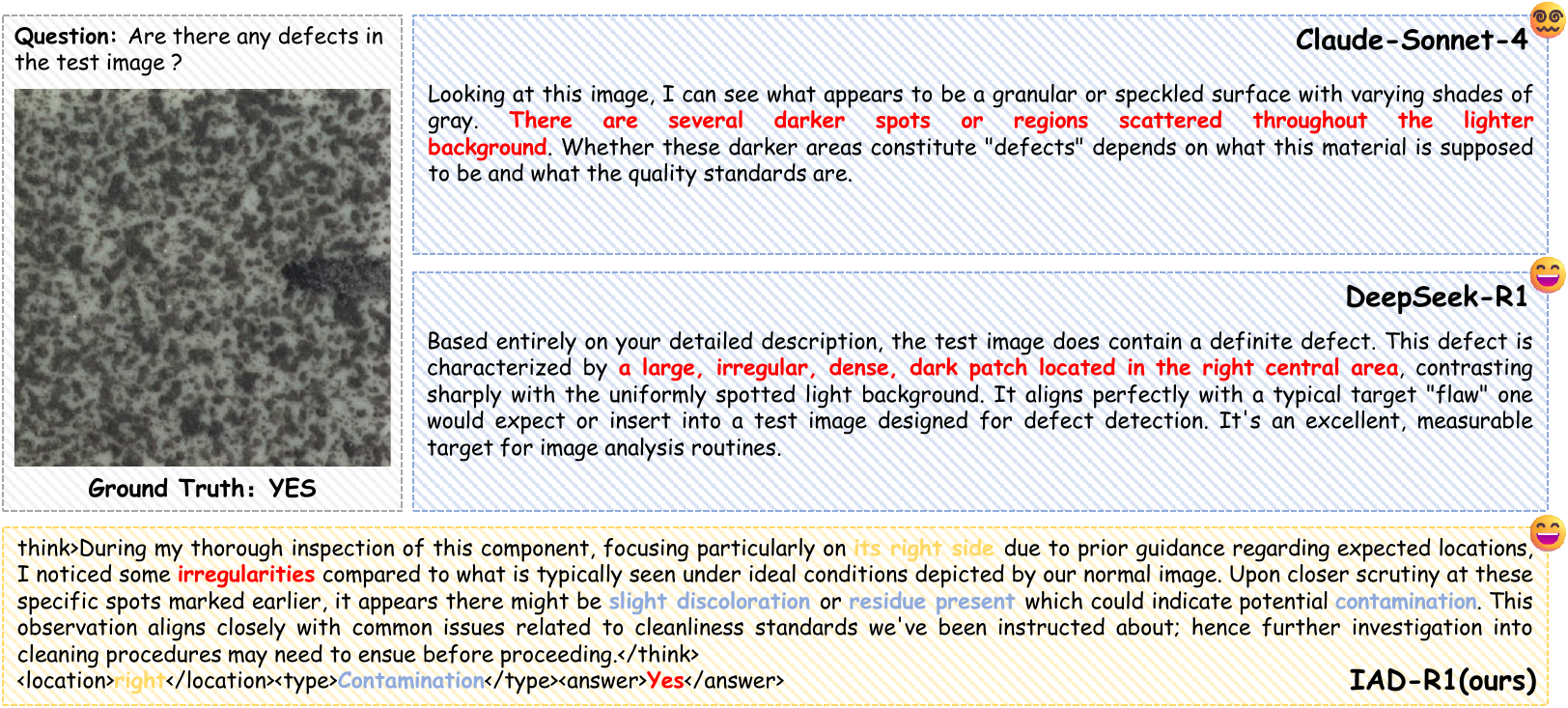}
    \caption{Different models' responses to a tile defect detection task.}
    \label{vis_tile}
\end{figure*}

\begin{figure*}[htbp]
    \centering
    \includegraphics[width=1\linewidth]{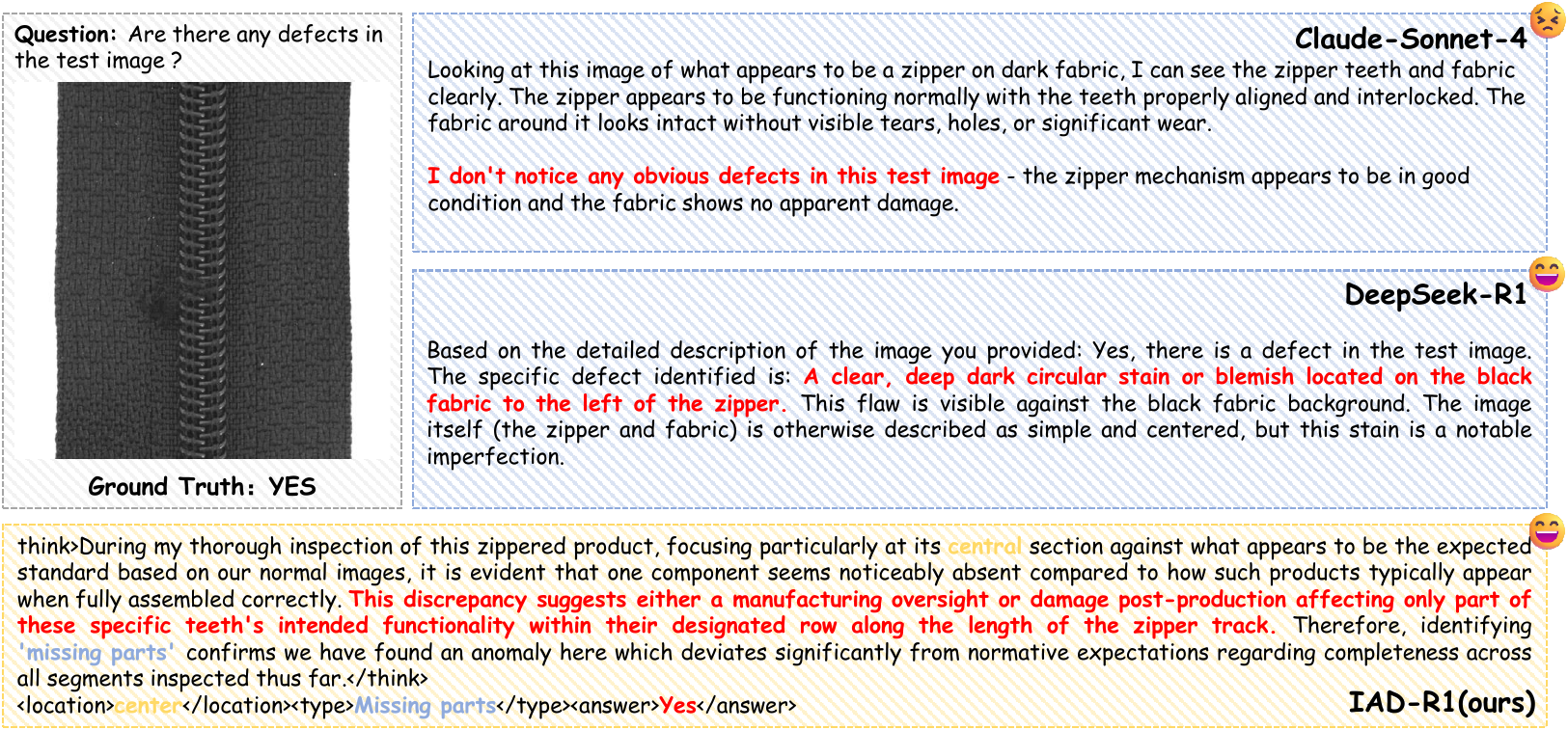}
    \caption{Different models' responses to a zipper defect detection task.}
    \label{vis_zipper}
\end{figure*}

\begin{figure*}[htbp]
    \centering
    \includegraphics[width=1\linewidth]{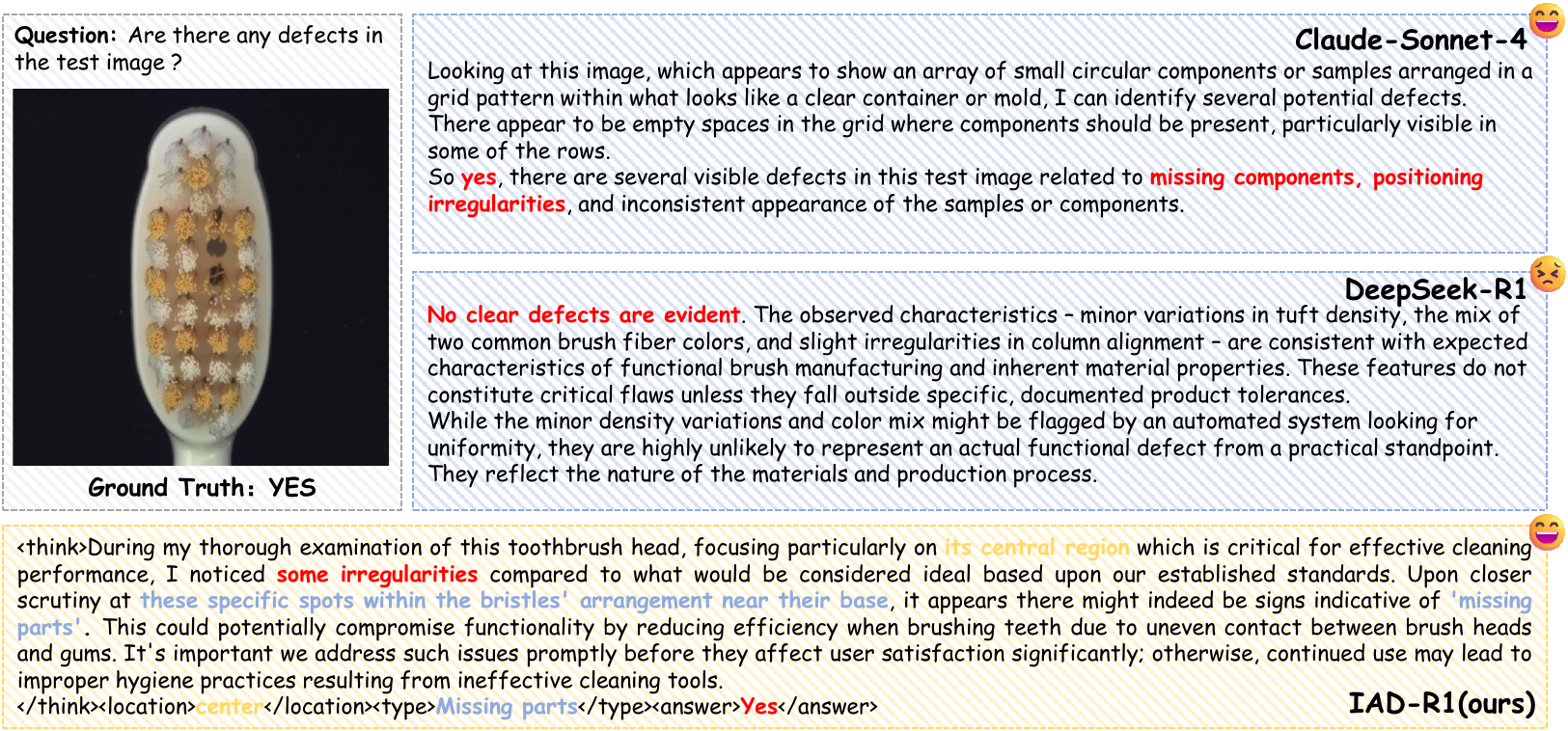}
    \caption{Different models' responses to a toothbrush defect detection task.}
    \label{vis_toothbrush}
\end{figure*}

\begin{figure*}[htbp]
    \centering
    \includegraphics[width=1\linewidth]{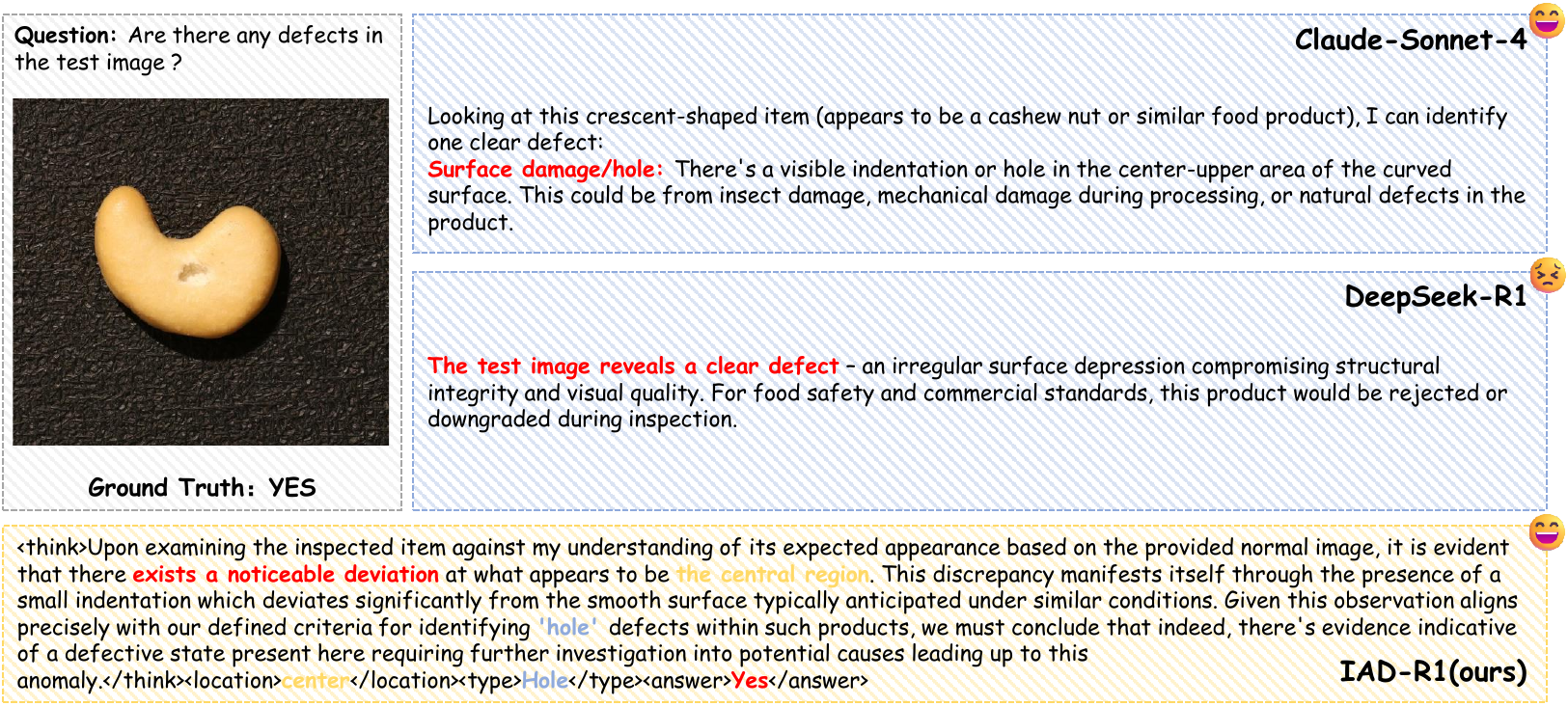}
    \caption{Different models' responses to a cashew defect detection task.}
    \label{vis_cashew}
\end{figure*}

\begin{figure*}[htbp]
    \centering
    \includegraphics[width=1\linewidth]{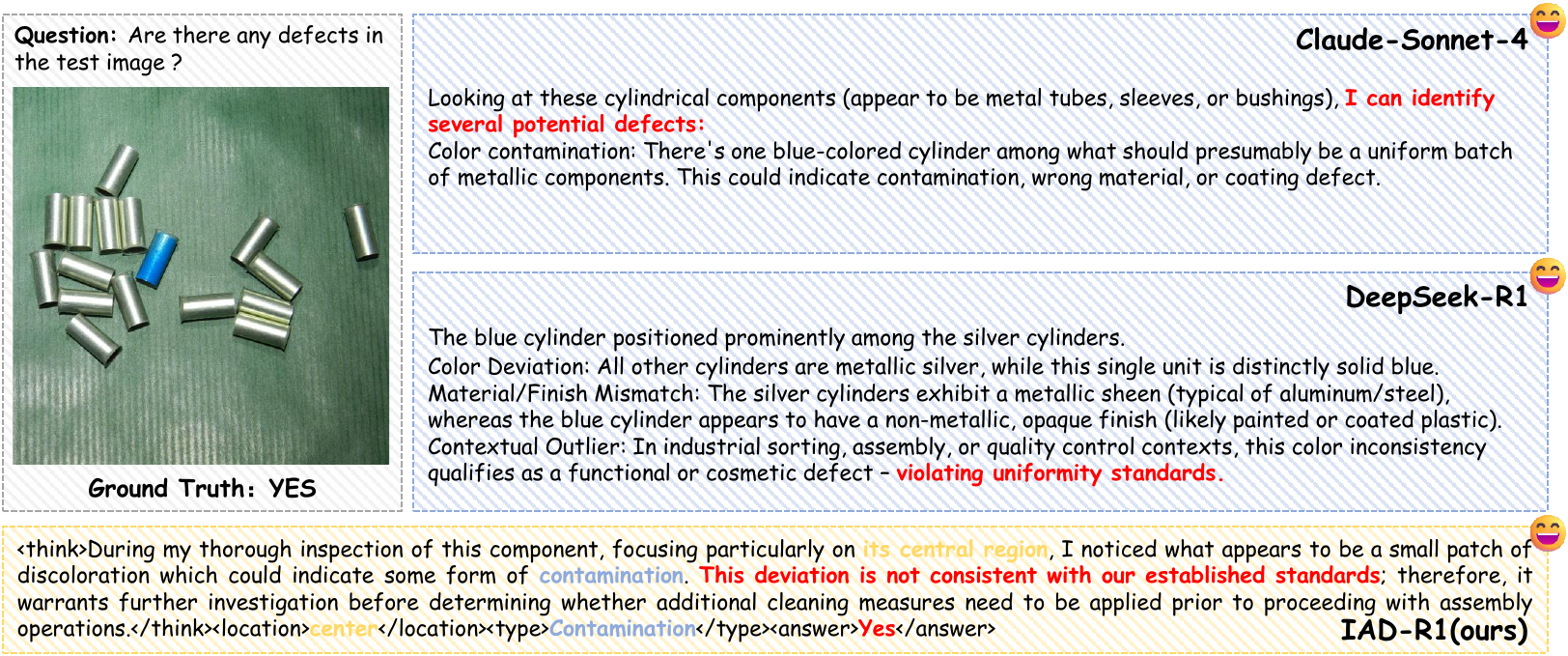}
    \caption{Different models' responses to a tube defect detection task.}
    \label{vis_tube}
\end{figure*}

\begin{figure*}[htbp]
    \centering
    \includegraphics[width=1\linewidth]{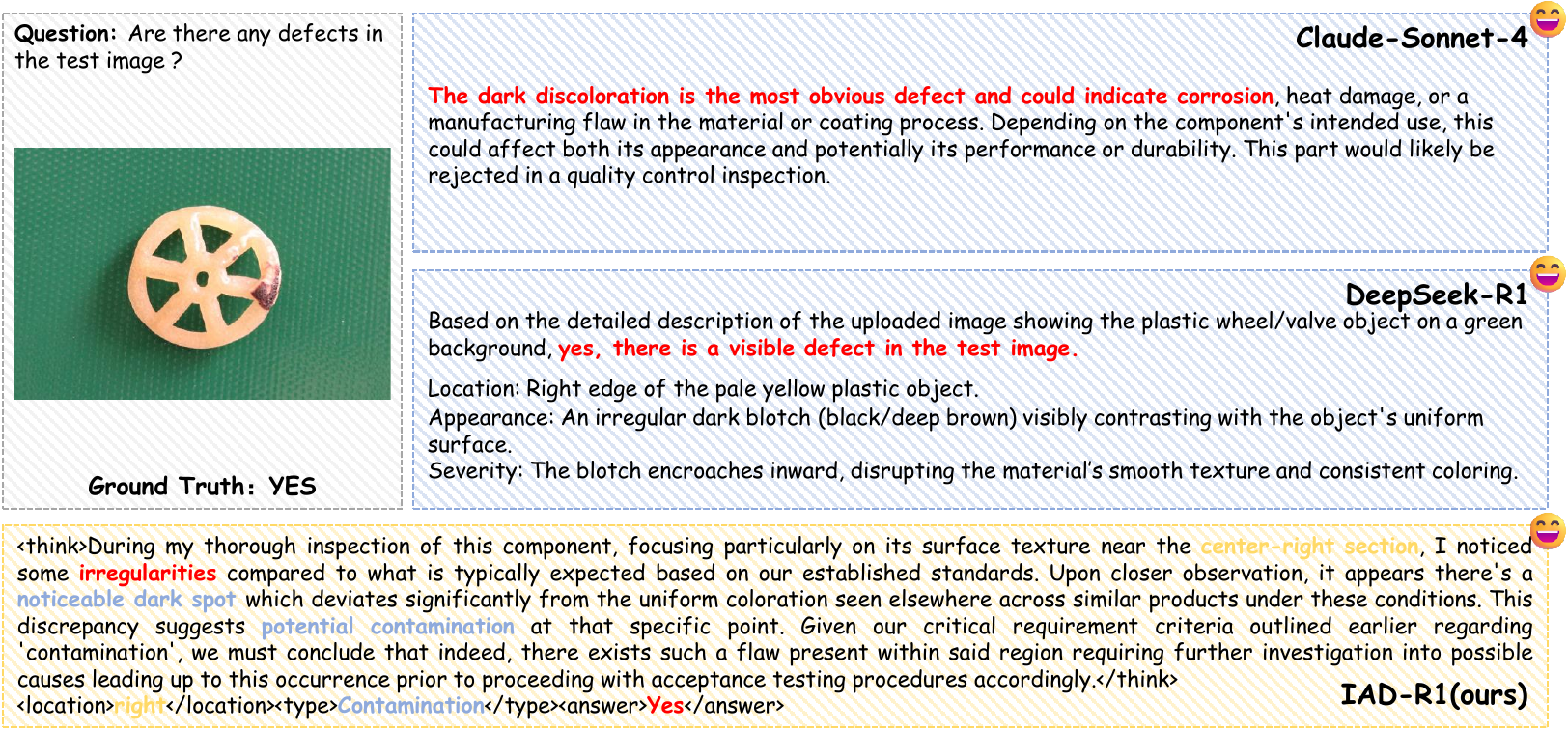}
    \caption{Different models' responses to a fryum defect detection task.}
    \label{vis_fryum}
\end{figure*}

\begin{figure*}[htbp]
    \centering
    \includegraphics[width=1\linewidth]{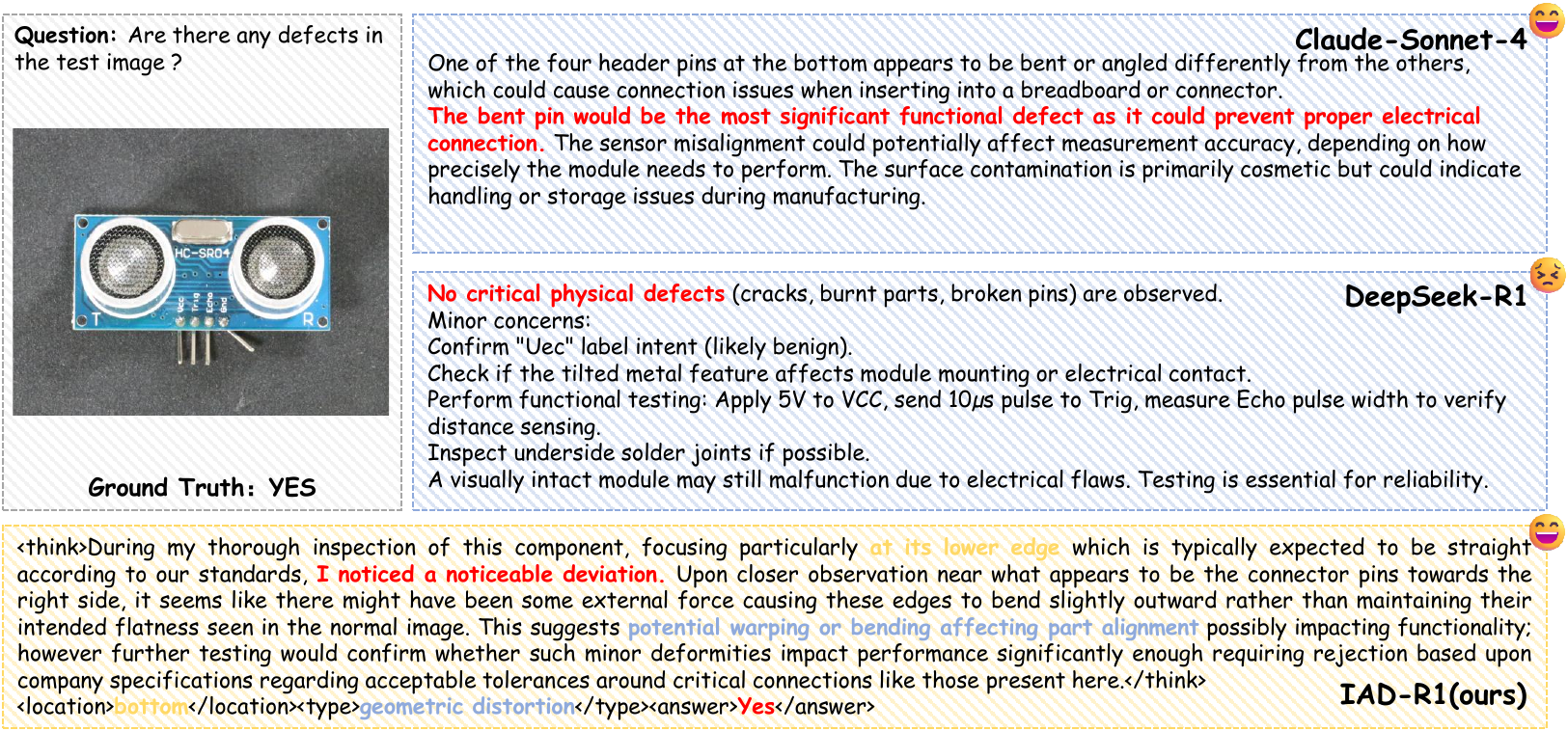}
    \caption{Different models' responses to a pcb defect detection task.}
    \label{vis_pcb_1}
\end{figure*}

\begin{figure*}[htbp]
    \centering
    \includegraphics[width=1\linewidth]{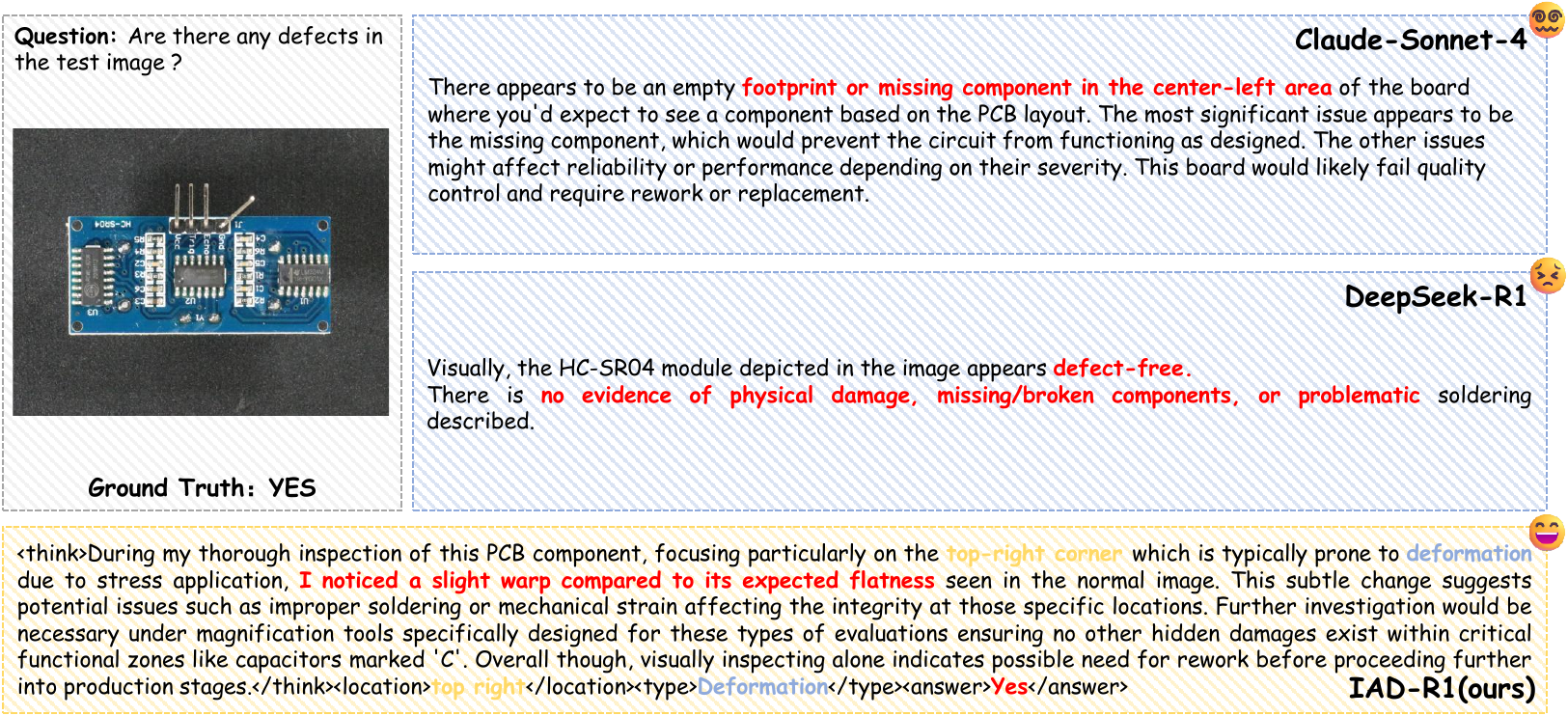}
    \caption{Different models' responses to a pcb defect detection task.}
    \label{vis_pcb_2}
\end{figure*}

\begin{figure*}[htbp]
    \centering
    \includegraphics[width=1\linewidth]{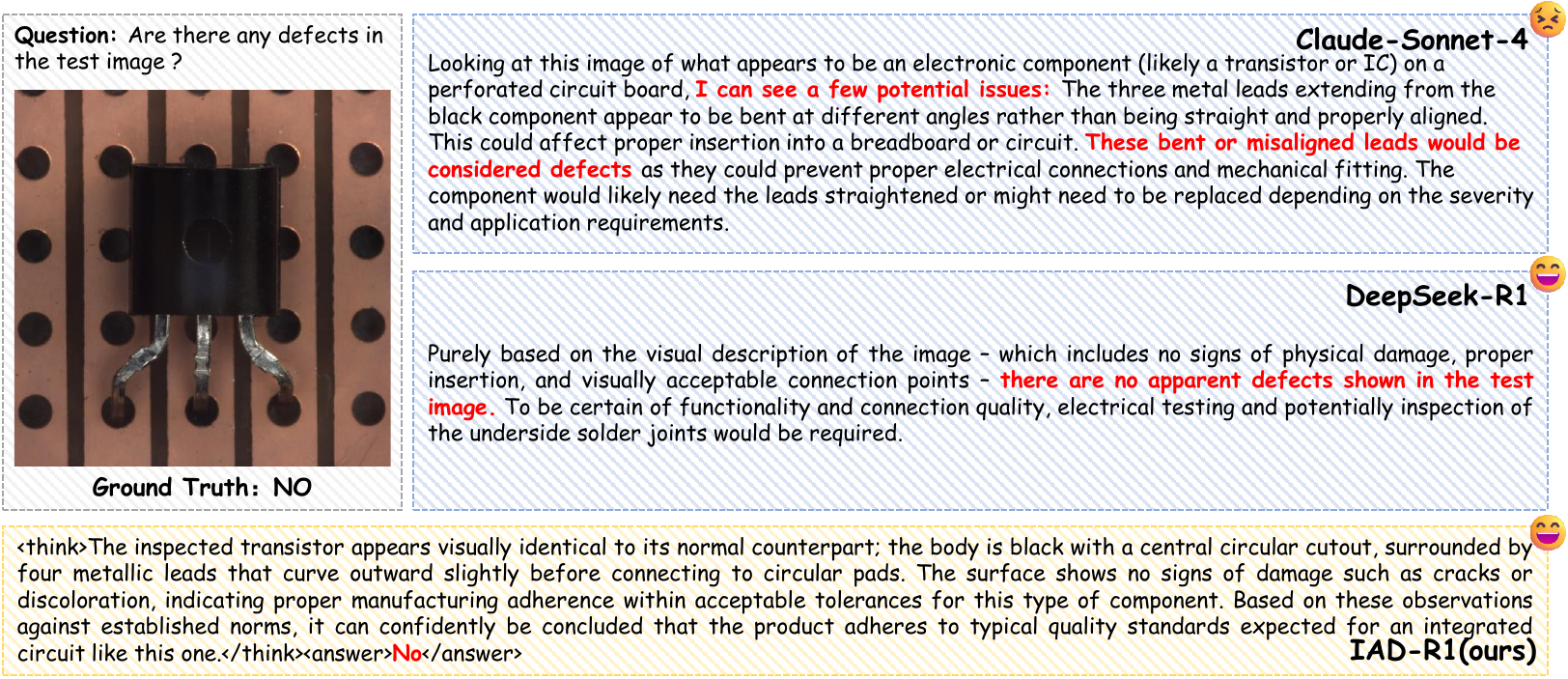}
    \caption{Different models' responses to a transistor defect detection task.}
    \label{vis_transistor}
\end{figure*}

\begin{figure*}[htbp]
    \centering
    \includegraphics[width=1\linewidth]{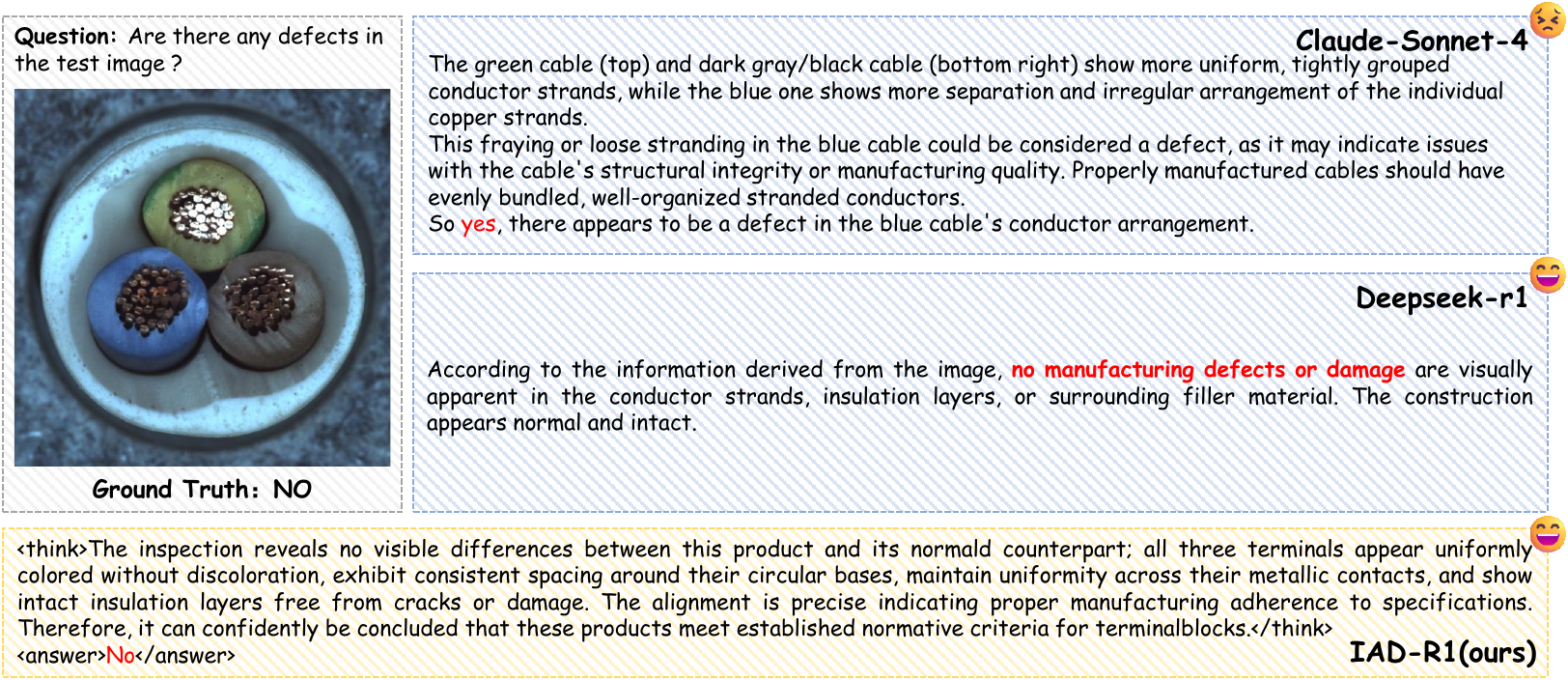}
    \caption{Different models' responses to a cable defect detection task.}
    \label{vis_cable}
\end{figure*}

Figures \ref{vis_screw}-\ref{vis_cable} show more model output comparisons.
\subsubsection{C.7. How about using base model as initial policy model?}
Although the R1-style paradigm~\cite{deepseek-r1} (i.e., using supervised fine-tuning as the initial policy model) has been proven effective in domains such as mathematical reasoning and code generation, its applicability to industrial anomaly detection remains to be validated. To explore the impact of different initial policy models on final performance, we conducted comparative experiments by directly applying the SC-GRPO strategy on base models without the PA-SFT stage.

Table \ref{initial_policy_model} reveals a significant phenomenon: most models exhibit severe reward bias problems during training, where models tend to bias their predictions toward a single category in pursuit of higher rewards. This phenomenon is particularly pronounced in class-imbalanced anomaly detection tasks. Such bias not only reduces the model's generalization capability but also violates the fundamental requirement of anomaly detection tasks to accurately identify both normal and abnormal samples.
In contrast, models using PA-SFT processed models as initial policy models demonstrate significantly different learning behaviors. The PA-SFT stage enables models to master structured output formats and fundamental anomaly perception capabilities, providing a stable initial policy model for subsequent reinforcement learning training and effectively preventing the occurrence of reward bias problems.
% 更多的可视化结果？训练曲线

\begin{figure}[!htbp]
    \centering
    \includegraphics[width=1\linewidth]{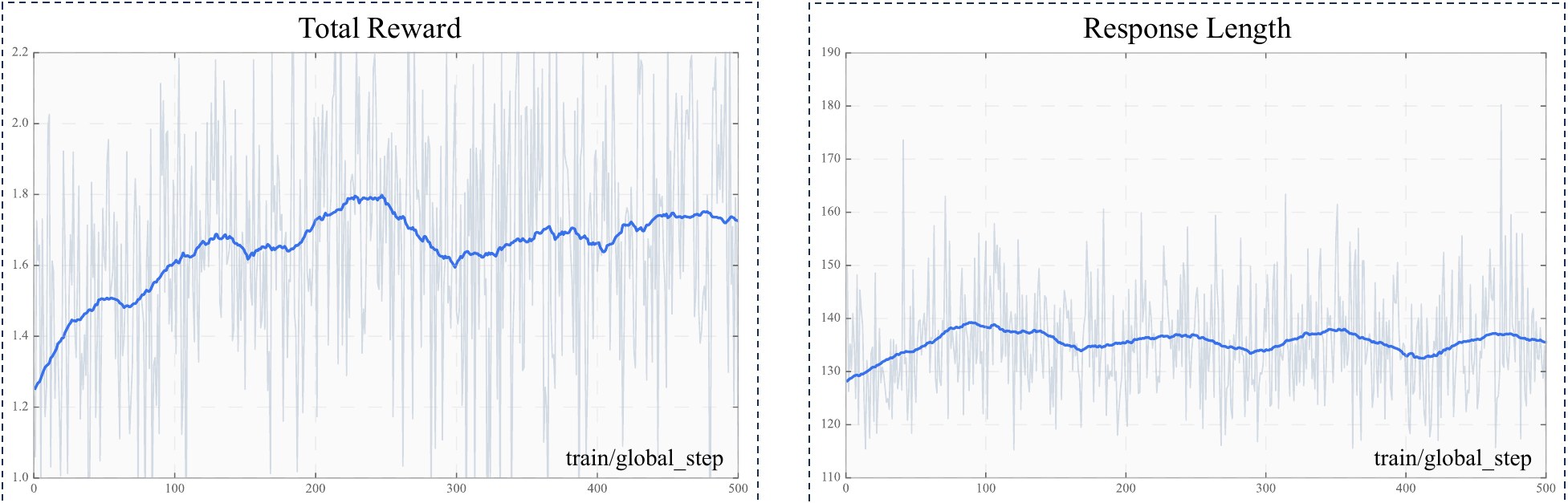}
    \caption{Training dynamics of Accuracy Reward and Response Length during SC-GRPO training process on LLaVA-1.5-7B. The left panel shows the reward progression over training steps, while the right panel displays the corresponding response length variations.}
    \label{curve_llava1.5}
\end{figure}
\subsubsection{C.8. Training Dynamics}
Figure \ref{curve_llava1.5} illustrates the learning dynamics during SC-GRPO training. The left panel demonstrates that the total reward exhibits a steady increase and converges to stability throughout the training process, indicating that the model can effectively learn anomaly detection strategies. The right panel shows that the response length remains within a reasonable range, demonstrating the stability of model outputs. These training dynamics characteristics reflect the favorable convergence properties and training stability of the SC-GRPO algorithm on industrial anomaly detection tasks.
\subsubsection{C.9. Model Promotion}
\begin{figure*}[htbp]
    \centering
    \includegraphics[width=1\linewidth]{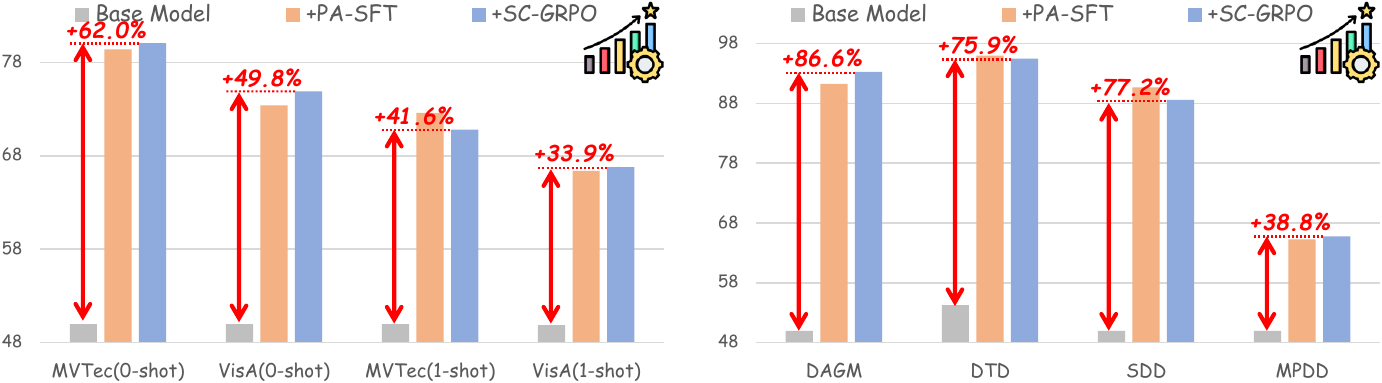}
    \caption{Performance comparison of different training strategies across multiple benchmarks on LLaVa-OneVision-SI-0.5B. Base Model (gray), +PA-SFT (orange), and +SC-GRPO (blue) are evaluated on MVTec, VisA, DAGM, DTD, SDD, and MPDD datasets. Red arrows and percentages indicate improvements achieved by IAD-R1 training over the base model.}
    \label{promotion_llava_0.5b}
\end{figure*}

\begin{figure*}[htbp]
    \centering
    \includegraphics[width=1\linewidth]{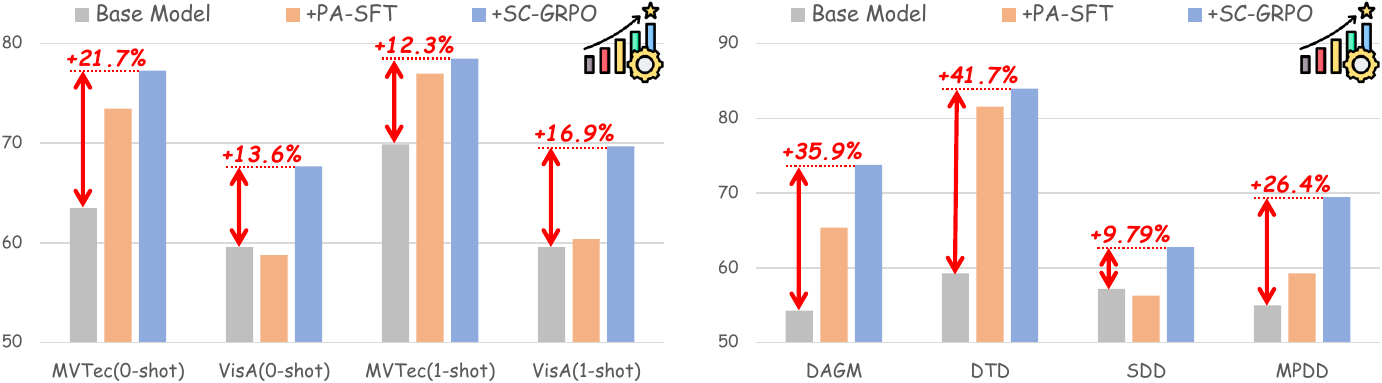}
    \caption{Performance comparison of different training strategies across multiple benchmarks on Qwen2-VL-Instruct-2B. Base Model (gray), +PA-SFT (orange), and +SC-GRPO (blue) are evaluated on MVTec, VisA, DAGM, DTD, SDD, and MPDD datasets. Red arrows and percentages indicate improvements achieved by IAD-R1 training over the base model.}
    \label{promotion_qwen2b}
\end{figure*}

\begin{figure*}[htbp]
    \centering
    \includegraphics[width=1\linewidth]{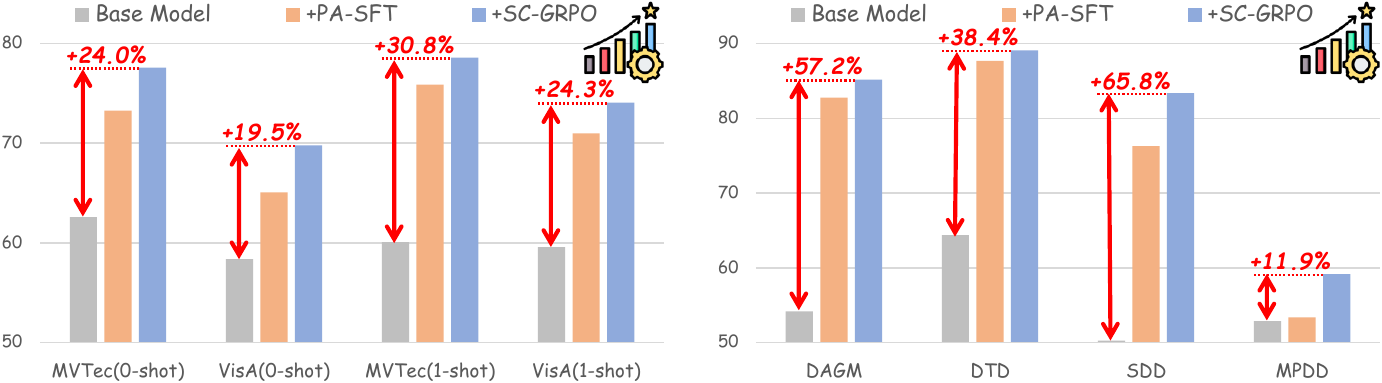}
    \caption{Performance comparison of different training strategies across multiple benchmarks on Qwen2.5-VL-Instruct-3B. Base Model (gray), +PA-SFT (orange), and +SC-GRPO (blue) are evaluated on MVTec, VisA, DAGM, DTD, SDD, and MPDD datasets. Red arrows and percentages indicate improvements achieved by IAD-R1 training over the base model.}
    \label{promotion_qwen3b}
\end{figure*}

\begin{figure*}[htbp]
    \centering
    \includegraphics[width=1\linewidth]{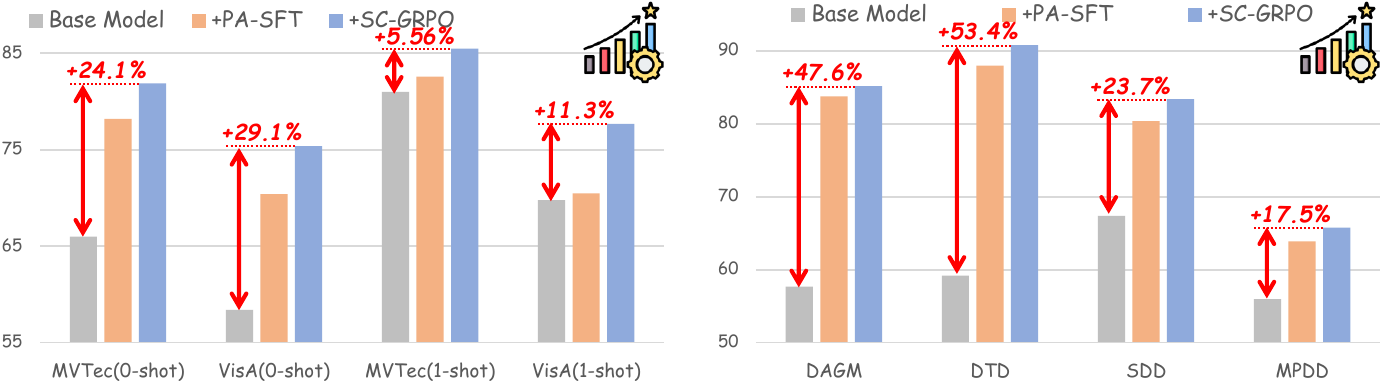}
    \caption{Performance comparison of different training strategies across multiple benchmarks on Qwen2.5-VL-Instruct-7B. Base Model (gray), +PA-SFT (orange), and +SC-GRPO (blue) are evaluated on MVTec, VisA, DAGM, DTD, SDD, and MPDD datasets. Red arrows and percentages indicate improvements achieved by IAD-R1 training over the base model.}
    \label{promotion_qwen7b}
\end{figure*}

\begin{figure*}[htbp]
    \centering
    \includegraphics[width=1\linewidth]{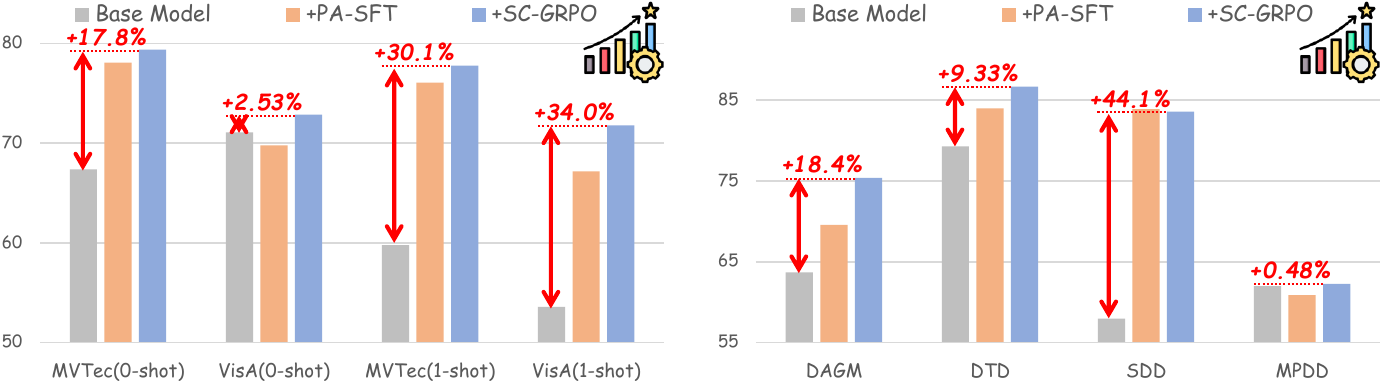}
    \caption{Performance comparison of different training strategies across multiple benchmarks on LLaVa-1.5-7B. Base Model (gray), +PA-SFT (orange), and +SC-GRPO (blue) are evaluated on MVTec, VisA, DAGM, DTD, SDD, and MPDD datasets. Red arrows and percentages indicate improvements achieved by IAD-R1 training over the base model.}
    \label{promotion_llava-1.5}
\end{figure*}

\begin{figure*}[htbp]
    \centering
    \includegraphics[width=1\linewidth]{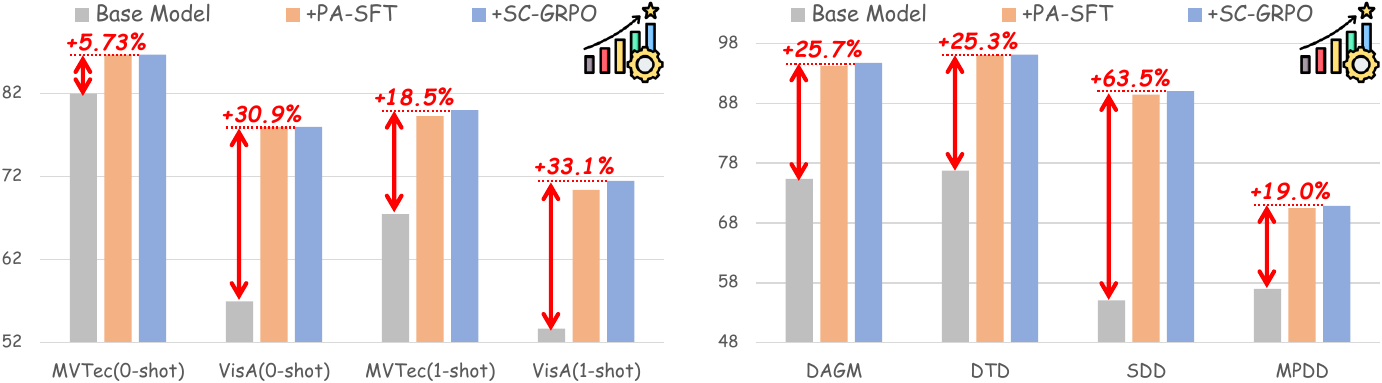}
    \caption{Performance comparison of different training strategies across multiple benchmarks on LLaVa-OneVision-SI-7B. Base Model (gray), +PA-SFT (orange), and +SC-GRPO (blue) are evaluated on MVTec, VisA, DAGM, DTD, SDD, and MPDD datasets. Red arrows and percentages indicate improvements achieved by IAD-R1 training over the base model.}
    \label{promotion_llava-si-7b}
\end{figure*}

\begin{figure*}[htbp]
    \centering
    \includegraphics[width=1\linewidth]{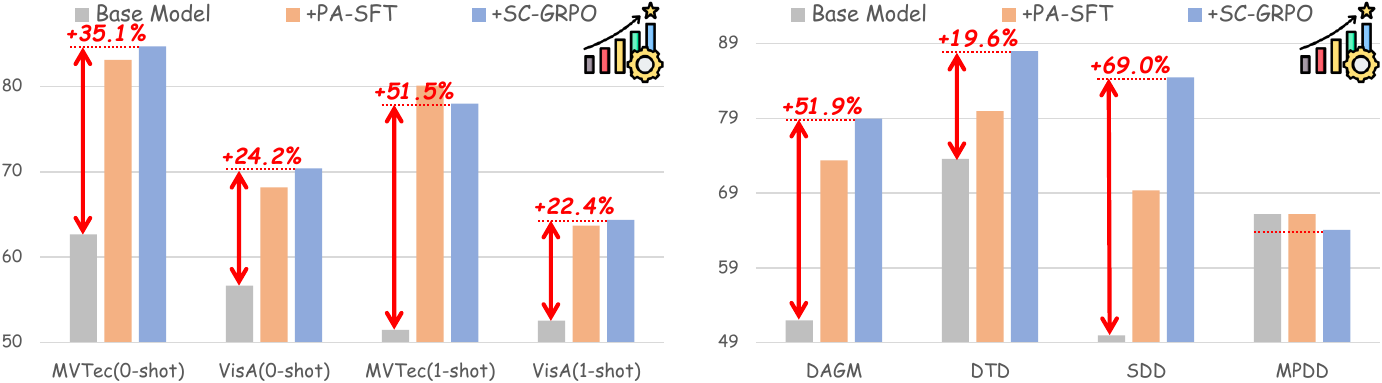}
    \caption{Performance comparison of different training strategies across multiple benchmarks on LLaVA-1.6-8B. Base Model (gray), +PA-SFT (orange), and +SC-GRPO (blue) are evaluated on MVTec, VisA, DAGM, DTD, SDD, and MPDD datasets. Red arrows and percentages indicate improvements achieved by IAD-R1 training over the base model.}
    \label{promotion_llava8b}
\end{figure*}

% 这个地方描述+PA-SFT的含义 +SC-GRPO的含义，指出提升是相对提升（相对base）
To visualize the progressive improvements achieved by our two-stage training approach, we present comprehensive performance comparisons across different model architectures and datasets. The visualization employs a systematic notation to illustrate the incremental benefits of each training stage: ``Base Model" (gray) represents the original pre-trained model without any anomaly detection fine-tuning; ``+PA-SFT" (orange) indicates the model after applying our perception activation supervised fine-tuning strategy on the Base Model; ``+SC-GRPO" (blue) represents the final model after further applying our structured control group relative policy optimization on the PA-SFT fine-tuned model. As demonstrated in Figures \ref{promotion_llava_0.5b}-\ref{promotion_llava8b}, The red arrows and percentages indicate the relative improvements achieved by the complete IAD-R1 training pipeline compared to the base model.
\subsection{D. Discussion and Future Work} \label{appendix_D}
\subsubsection{D.1. Limitations}
Although Expert-AD provides high-quality CoT data, its dataset scale is relatively limited, which may affect the model's generalization capability across broader industrial scenarios.
\subsubsection{D.2. Future Work}
We plan to further improve and extend the IAD-R1 framework from multiple perspectives. First, in terms of data, we will collect more real industrial scenario images and construct corresponding high-quality CoT responses to enhance dataset scale and diversity, improving model generalization across different industrial environments. Second, we will apply IAD-R1 to more backbone models with different architectures and parameter scales, including the latest open-source VLMs, to comprehensively validate the universality and effectiveness of the IAD-R1 framework. Additionally, we plan to incorporate more anomaly detection datasets into the testing benchmark, covering broader industrial domains and anomaly types for comprehensive model performance evaluation. Finally, we will continuously track and evaluate newly released open-source models and commercial APIs to ensure the IAD-R1 framework stays synchronized with cutting-edge technologies and continuously validate its relative advantages.

\newpage
\begin{table*}[htbp]
\centering
\small
\begin{tabular}{ccccccccccc}
\toprule
\multirow{2}{*}{\textbf{Model}} & \multirow{2}{*}{\textbf{Parameter}} & \multirow{2}{*}{\textbf{Data}} & \multicolumn{6}{c}{\textbf{0-shot}} & \multicolumn{2}{c}{\textbf{1-shot}} \\
\cmidrule(lr){4-9} \cmidrule(lr){10-11}
& & & \textbf{MVTec} & \textbf{VisA} & \textbf{DAGM} & \textbf{DTD} & \textbf{MPDD} & \textbf{SDD} & \textbf{MVTec} & \textbf{VisA} \\
\midrule
\multirow{3}{*}{LLaVA-OneVision-SI} & \multirow{3}{*}{0.5B} 
& Base & 50.0 & 50.0 & 50.0 & 54.3 & 50.0 & 50.0 & 50.0 & 49.9 \\
& & Original & 47.9 & 50.0 & 50.0 & 50.9 & \textbf{56.5} & 50.0 & 47.9 & \textbf{52.0} \\
& & Expert-AD & \textbf{79.4} & \textbf{73.4} & \textbf{91.3} & \textbf{96.0} & 65.3 & \textbf{96.0} & \textbf{72.6} & \textbf{66.4} \\
\multirow{3}{*}{Qwen2-VL-Instruct} & \multirow{3}{*}{2B} 
& Base & 63.5 & \textbf{59.6} & 54.3 & 59.3 & \textbf{55.0} & \textbf{57.2} & \textbf{69.9} & \textbf{59.6} \\
& & Original & 54.0 & 55.1 & 53.5 & \textbf{67.3} & 48.3 & 50.0 & 54.4 & 57.2 \\
& & Expert-AD & \textbf{73.5} & 58.8 & \textbf{65.4} & \textbf{81.6} & \textbf{59.3} & 56.3 & \textbf{77.0} & 60.4 \\
\multirow{3}{*}{Qwen2.5-VL-Instruct} & \multirow{3}{*}{3B} 
& Base & 62.6 & 58.4 & 54.2 & 64.4 & 52.9 & 50.3 & 60.1 & 59.6 \\
& & Original & 53.7 & 64.9 & 66.0 & 77.2 & 58.2 & \textbf{74.5} & 53.5 & 65.4 \\
& & Expert-AD & \textbf{73.3} & \textbf{65.1} & \textbf{82.8} & \textbf{87.7} & \textbf{58.2} & 76.3 & \textbf{75.9} & \textbf{71.0} \\
\multirow{3}{*}{LLaVA-OneVision-SI} & \multirow{3}{*}{7B} 
& Base & 82.0 & 59.6 & 75.4 & \textbf{76.8} & 57.0 & 55.1 & \textbf{67.5} & 53.7 \\
& & Original & 64.0 & 71.7 & \textbf{78.4} & 66.2 & 58.8 & \textbf{71.1} & 60.6 & 63.9 \\
& & Expert-AD & \textbf{86.7} & \textbf{78.0} & \textbf{94.3} & \textbf{96.1} & \textbf{70.5} & \textbf{89.5} & 79.3 & \textbf{70.4} \\
\bottomrule
\end{tabular}%
\caption{Detailed results of different data preparation in PA-SFT.}
\label{appendix_ablation_sft}
\end{table*}

\begin{table*}[htbp]
\centering
\small
\begin{tabular}{ccccccccccc}
\toprule
\multirow{2}{*}{\textbf{Model}} & \multirow{2}{*}{\textbf{Parameter}} & \multirow{2}{*}{\textbf{Strategy}} & \multicolumn{6}{c}{\textbf{0-shot}} & \multicolumn{2}{c}{\textbf{1-shot}} \\
\cmidrule(lr){4-9} \cmidrule(lr){10-11}
& & & \textbf{MVTec} & \textbf{VisA} & \textbf{DAGM} & \textbf{DTD} & \textbf{MPDD} & \textbf{SDD} & \textbf{MVTec} & \textbf{VisA} \\
\midrule
\multirow{3}{*}{LLaVA-OneVision-SI} & \multirow{3}{*}{0.5B} 
& Fine-tuning & 79.4 & 73.4 & 91.3 & \textbf{96.0} & 65.3 & \textbf{90.7} & \textbf{72.6} & \textbf{66.4} \\
& & Original & 76.7 & 67.4 & 89.9 & 93.8 & \textbf{66.4} & 74.5 & 60.8 & 57.2 \\
& & SC-GRPO & \textbf{81.0} & \textbf{74.9} & \textbf{93.3} & 95.5 & \textbf{69.4} & 88.6 & 70.8 & 66.8 \\
\multirow{3}{*}{Qwen2-VL-Instruct} & \multirow{3}{*}{2B} 
& Fine-tuning & 73.5 & 58.8 & \textbf{65.4} & \textbf{81.6} & 59.3 & \textbf{56.3} & \textbf{77.0} & 60.4 \\
& & Original & 75.7 & 58.7 & 54.3 & 68.6 & \textbf{69.6} & 53.7 & 75.8 & 62.2 \\
& & SC-GRPO & \textbf{77.3} & \textbf{67.7} & 73.8 & 84.0 & \textbf{69.5} & 62.8 & 78.5 & \textbf{69.7} \\
\multirow{3}{*}{Qwen2.5-VL-Instruct} & \multirow{3}{*}{3B} 
& Fine-tuning & 73.3 & 65.1 & 82.8 & \textbf{87.7} & 53.4 & 76.3 & 75.9 & 71.0 \\
& & Original & 72.7 & 59.0 & 72.5 & 81.0 & \textbf{55.5} & \textbf{79.3} & \textbf{76.8} & 67.0 \\
& & SC-GRPO & \textbf{77.6} & \textbf{69.8} & \textbf{85.2} & \textbf{89.1} & \textbf{59.2} & \textbf{83.4} & \textbf{78.6} & \textbf{74.1} \\
\multirow{3}{*}{LLaVA-OneVision-SI} & \multirow{3}{*}{7B} 
& Fine-tuning & \textbf{86.7} & \textbf{78.0} & 94.3 & 96.1 & 70.5 & 89.5 & 79.3 & 70.4 \\
& & Original & 82.1 & 70.1 & 84.4 & 88.8 & 66.3 & 79.3 & 69.1 & 62.4 \\
& & SC-GRPO & \textbf{86.7} & \textbf{78.0} & \textbf{94.8} & \textbf{96.2} & \textbf{70.9} & \textbf{90.1} & \textbf{80.0} & \textbf{71.5} \\
\bottomrule
\end{tabular}%
\caption{Detailed results of different reward strategy in SC-GRPO.}
\label{appendix_ablation_grpo}
\end{table*}

%模型输出对比结果

\end{document}